\documentclass{article} 
\usepackage{iclr2027_conference,times}

\usepackage{amsmath,amsfonts,bm}

\def\eqref#1{equation~\ref{#1}}

\def\1{\bm{1}}

\DeclareMathAlphabet{\mathsfit}{\encodingdefault}{\sfdefault}{m}{sl}
\SetMathAlphabet{\mathsfit}{bold}{\encodingdefault}{\sfdefault}{bx}{n}

\usepackage{hyperref}
\usepackage{url}

\usepackage[utf8]{inputenc} 
\usepackage[T1]{fontenc}    
\usepackage{url}            
\usepackage{booktabs}       
\usepackage{amsmath}
\usepackage{amsfonts}       
\usepackage{nicefrac}       
\usepackage{microtype}      
\usepackage{xcolor}         
\usepackage{comment}
\usepackage{graphicx}
\usepackage{multirow}
\usepackage{enumitem}
\usepackage{tabularx}
\usepackage{pifont}
\usepackage{tikz}
\usetikzlibrary{positioning}
\usepackage{siunitx}
\usepackage{cleveref}
\usepackage{subcaption}

\newcommand{\todo}[1]{\textbf{\color{red}[TODO: #1]}}

\usepackage{soul}
\setuldepth{foobar}

\definecolor{green}{RGB}{25, 169, 116}
\definecolor{red}{RGB}{255, 65, 54}
\newcommand{\cmark}{\textcolor{green}{\ding{51}}}
\newcommand{\xmark}{\textcolor{red}{\ding{55}}}
\definecolor{blue}{RGB}{0, 18, 115}
\definecolor{cyan}{RGB}{10, 147, 150}
\definecolor{sea}{RGB}{148, 210, 189}

\usepackage[accsupp]{axessibility}  

\title{Towards Open-Ended Visual Scientific Discovery with Sparse Autoencoders} 

\iclrfinaltrue

\author{Jacob Beattie$^{1*}$, Samuel Stevens$^{1}$, Neil Rosser$^{2}$, Yu Su$^{1}$, Tanya Berger-Wolf$^{1}$ \\
$^1$The Ohio State University, $^2$The University of Miami\\
$^*$\texttt{beattie.74@osu.edu}
}

\makeatletter
\def\@maketitle{\vbox{\hsize\textwidth
{\LARGE\sc \@title\par}
\def\And{\end{tabular}\hfil\linebreak[0]\hfil
        \begin{tabular}[t]{l}\bf\rule{\z@}{24pt}\ignorespaces}%
\def\AND{\end{tabular}\hfil\linebreak[4]\hfil
        \begin{tabular}[t]{l}\bf\rule{\z@}{24pt}\ignorespaces}%
\begin{tabular}[t]{l}\bf\rule{\z@}{24pt}\@author\end{tabular}%
\vskip 0.3in minus 0.1in}}
\makeatother

\begin{document}

\maketitle

\begin{abstract}
Foundation models in several scientific domains, including visual domains, learn representations that capture complex semantics from their respective fields.
Despite this, most existing applications focus on pre-specified concepts or targets, excelling at confirmation but not open-ended discovery of unknown patterns. 
We study whether sparse autoencoders (SAEs) with heuristic feature ranking can provide a practical methodology for surfacing candidate visual features from vision foundation models for scientific discovery without concept-specific supervision.
We evaluate this approach in three stages: (1) general-purpose concept rediscovery, (2) domain-specific concept rediscovery, and (3) question-driven feature ranking.
On ADE20K, SAEs outperform baseline methods ($k$-means, PCA, SemiNMF) across most downstream metrics for concept rediscovery.
On FishVista, the same approach recovers fine-grained fish anatomical features without concept supervision.
Finally, on \textit{Heliconius} butterflies, a heuristic based on the Mann-Whitney $U$ statistic consistently surfaces features corresponding to real diagnostic keys between mimic subspecies.
Together, these results show that SAEs with heuristic ranking can efficiently learn and surface semantic structure in vision foundation model representations.
We view this as a methodological prerequisite for open-ended visual scientific discovery: the present experiments validate feature surfacing and ranking through rediscovery, while prospective discovery of previously unknown scientific phenomena remains future work.

\end{abstract}

\section{Introduction}

Scientific data now span hundreds of petabytes across domains, far outpacing what any lab can annotate or manually inspect.\footnote{For example, the NIH Sequence Read Archive exposes roughly \num{47} PB of public sequencing data, the European Nucleotide Archive provides roughly \num{63} PB of immediately downloadable data, NASA Earthdata (EOSDIS) hosts more than \num{120} PB of open Earth science data, and a single reanalysis, ERA5, is on the order of \num{5} PB.}
Large neural networks trained on these datasets cover proteins \citep{abramson2024alphafold3}, weather \citep{lam2023graphcast,price2025gencast}, genomics \citep{cui2024scgpt,hayes2025esm3}, geospatial \citep{reed2023scale}, and ecology \citep{gu2025bioclip2}.
These models achieve strong predictive performance across a variety of tasks by learning compressed representations of the underlying data distribution's structure.

\begin{figure*}[t]
    \centering
    \includegraphics[width=\linewidth]{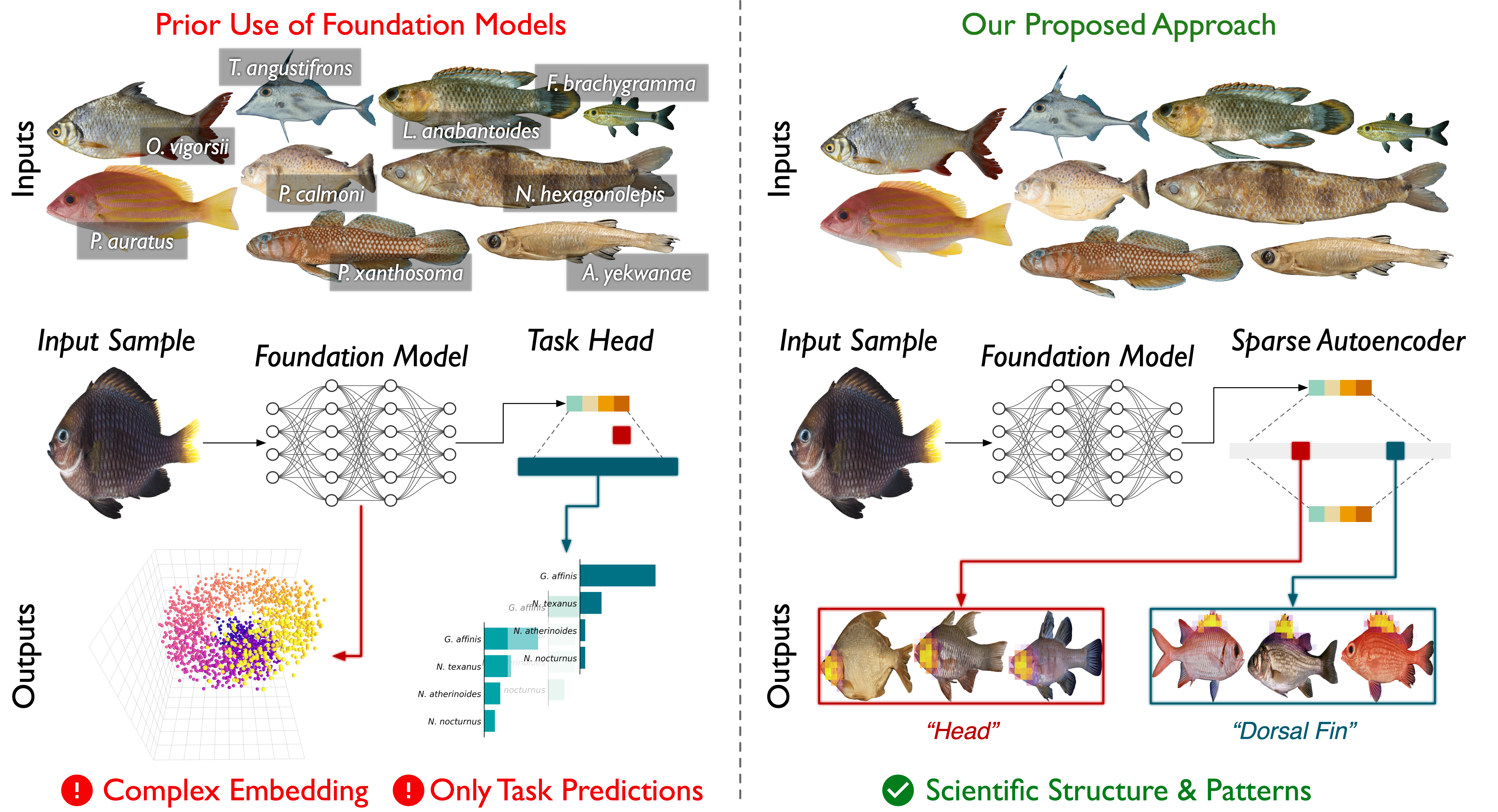}
    \caption{\textit{Validating an instrument for open-ended feature discovery.} 
    \textbf{Left:} Typical use of a foundation model in science: an input image is passed through a pretrained encoder and a task-specific head to yield class scores; the dense representation remains an opaque embedding, so unnamed factors are inaccessible. 
    \textbf{Right:} Our procedure composes a foundation model with a sparse autoencoder, producing a library of interpretable semantic concepts with per-example activation maps. 
    We find that these concepts align with and localize anatomical parts (e.g., head, dorsal fin) without seeing part-of-body labels.
    The learned concepts are inherently quantitative, making them easy to rank and search using heuristics.
    }\label{fig:hook}
\end{figure*}

However, foundation models are currently limited to \emph{confirmatory} applications where the target concepts are pre-specified (see \Cref{fig:related-work}).
Foundation models are used as frozen feature extractors or fine-tuned for specific predictions (species classification \citep{vanhorn2021inat21}, cellular responses to perturbations \citep{roohani2025vcc}, 3-D protein structure \citep{haas2018continuous,pereira2021high}).
Because of this, the representations produced by foundation models are high-dimensional embeddings optimized for task performance.
While these representations enable accurate predictions, they do not directly provide reusable scientific measurements or expose the structure the model learned for systematic investigation.

\paragraph{Discovery vs. Confirmation}
Scientific discovery often involves establishing names and descriptions of specific, previously unexplained phenomena \citep{kuhn1970structure, poincare_science_2003}.
A morphological feature that distinguishes subspecies but is not formally cataloged, an unexplored correlation between climate variables, or a sequence motif in an understudied protein family: these cannot be found by methods that require specifying target concepts in advance.
Here, we focus on “open-ended feature discovery”: surfacing candidate visual features that are relevant to a specific scientific question, without specifying relevant concepts beforehand \citep{hughes2024open}.
Because genuinely novel scientific findings cannot be evaluated against ground truth, we evaluate this capability through controlled rediscovery of known concepts and traits.

Interpretability methods offer a route to systematically surface the semantic structure learned by foundation models.
While many existing interpretability methods excel at \textit{confirmation} \citep{alain2016understanding, kim2018tcav, selvaraju2017gradcam, sundararajan2017axiomatic,simonyan2013deep}, they cannot perform open-ended exploration to generate hypotheses (see \Cref{fig:related-work}).
In particular, methods for such open-ended discovery from \textit{vision} foundation model representations remain underdeveloped and untested.

\begin{figure}[t]
\centering
\begin{tikzpicture}[
  font=\footnotesize,
  card/.style={draw, rounded corners, align=left, inner sep=3pt, text width=0.47\columnwidth},
  good/.style={card, line width=0.9pt, fill=green!6},
  gap/.style={}
]

\node[card] (c1) {
  \textbf{Task Heads}\\
  \textit{Input:} \colorbox{red!15}{Labeled task dataset}\\
  \textit{Output:} Task predictions\\
  \textit{Example:} Species classification\\
  \textbf{Discovery:} \textcolor{red}{\xmark} Needs task labels
};

\node[card, right=3pt of c1] (c2) {
  \textbf{Linear Probes}\\
  \textit{Input:} \colorbox{red!15}{Labeled concept $A$ samples}\\
  \textit{Output:} Decodability of $A$\\
  \textit{Example:} ``Does DINOv3 see eyes?''\\
  \textbf{Discovery:} \textcolor{red}{\xmark} Pre-specified concepts
};

\node[card, below=3pt of c1] (c3) {
  \textbf{Concept Vectors}\\
  \textit{Input:} \colorbox{red!15}{Labeled concept $A$ samples}\\
  \textit{Output:} Score for concept $A$\\
  \textit{Example:} ``Find a striped direction''\\
  \textbf{Discovery:} \textcolor{red}{\xmark} Needs concept samples
};

\node[card, below=3pt of c2] (c4) {
  \textbf{Feature Viz.}\\
  \textit{Input:} Selected neuron\\
  \textit{Output:} \colorbox{red!15}{Synthetic images}\\
  \textit{Example:} ``DeepDream for neuron $j$''\\
  \textbf{Discovery:} \textcolor{red}{\xmark} Unstructured output
};

\node[card, below=3pt of c3] (c5) {
  \textbf{Prototypes}\\
  \textit{Input:} \colorbox{red!15}{Labeled task dataset}\\
  \textit{Output:} Task-tied prototypes\\
  \textit{Example:} ``This looks like concept $B$''\\
  \textbf{Discovery:} \textcolor{red}{\xmark} Task-specific
};

\node[good, below=3pt of c4] (c6) {
  \textbf{SAEs + Heuristics (Ours)}\\
  \textit{Input:} Unlabeled activations\\
  \textit{Output:} Quantified, ranked, interpretable features\\
  \textit{Example:} Discover new traits\\
  \textbf{Discovery:} \textcolor{green!50!black}{\cmark} Finds unknown patterns
};

\end{tikzpicture}
\caption{Comparison of methods for extracting knowledge from vision foundation models. 
    Most approaches require researchers to specify concepts or tasks in advance.}\label{fig:related-work}
\end{figure}

\paragraph{Our approach.}
We propose sparse autoencoders (SAEs) with heuristic ranking as a step towards open-ended visual feature discovery.
SAEs learn an overcomplete dictionary that reconstructs model activations through sparse linear combinations, with sparsity encouraging individual features to capture semantically coherent patterns \citep{bricken2023monosemanticity,cunningham2023sparse,templeton2024scaling,gao2025scaling}.
Unlike supervised methods, SAEs require no labels during training.
Each learned feature provides a per-example activation score together with its decoding direction and exemplar evidence, enabling systematic investigation of what patterns the model represents.
We then propose question-driven heuristics to rank and quickly surface relevant and possibly novel visual features. 
In short, we treat SAEs with heuristic feature ranking as a potential tool for open-ended visual feature discovery from foundation models and seek to rigorously evaluate that tool in controlled settings \citep{peng2025unknown}.

\paragraph{Evidence.}
We evaluate this approach across three experimental settings to determine its strengths and limitations\footnote{Full implementation can be found at \url{https://anonymous.4open.science/r/saev-25A1}}:
\begin{itemize}
    \item General-purpose rediscovery of segmentation concepts in ADE20K \citep{zhou2017ade20k}, comparing SAEs with other feature learning baselines across vision foundation models \citep{simeoni2025dinov3, bolya2025perception},
    \item Domain-specific rediscovery of segmentation concepts in FishVista \citep{mehrab2024fishvista,fishvistadataset}, comparing SAE variations and per-concept performance,
    \item Question-specific feature ranking in \textit{Heliconius} butterflies \citep{hdr_imageomics_institute_heliconius-collection_cambridge-butterfly_nodate}, testing feature ranking for efficient expert validation.
\end{itemize}

On ADE20K, we find that SAEs outperform baseline methods ($k$-means clustering, PCA, and SemiNMF) in concept alignment (\Cref{sec:ade20k,sec:ablations}), rediscovering known segmentation concepts without supervision.
On FishVista, we find that SAEs learn fine-grained anatomical features, providing measurable quantities that can be evaluated against biological annotations.
Finally, we find that a modified AUC-ROC score highly ranks known diagnostic traits across several mimic subspecies pairs, relying only on species labels (\Cref{sec:butterfly-results}).
Our results indicate that SAEs with heuristic ranking can learn and effectively surface semantic structure from vision foundation model representations.
We view this work as a prerequisite step to visual scientific discovery: establishing a methodology that can systematically recover and rank known structure from images, making search for truly novel scientific phenomena feasible.

\section{Related Work}

\paragraph{Foundation Models.}
Domain-specific foundation models capture rich structure and achieve state-of-the-art performance on their focal tasks, as shown in proteins \citep{abramson2024alphafold3,hayes2025esm3}, ecology \citep{stevens2024bioclip}, weather \citep{lam2023graphcast,price2025gencast}, and single-cell genomics \citep{cui2024scgpt}.
In both general and scientific domains, large-scale training leads to emergent abilities beyond the pre-training objective \citep{wei2022emergent,gu2025bioclip2}.
Theory and evidence suggest that large pretrained models internalize meaningful structure \citep{ha2018world,caron2021dinov1,lecun2022jepa,huh2024platonic}.
Despite this, most scientific models are benchmarked as \textbf{task-specific predictors} rather than sources of novel concepts.

\paragraph{Interpretability Methods.}
A large body of work explains predictions without exposing a global factorization of representations.
Saliency and attribution methods (e.g., gradients, Integrated Gradients, Grad-CAM) highlight input regions for \emph{individual} decisions and have known validity pitfalls \citep{simonyan2013deep,sundararajan2017axiomatic,selvaraju2017gradcam,adebayo2018sanity}.
Linear probes test separability of \emph{pre-specified} targets \citep{alain2016understanding,hewitt2019probes}; concept methods such as TCAV/ACE require exemplar sets and so test investigator-defined ideas \citep{kim2018tcav,ghorbani2019ace}.
Prototype models (ProtoPNet, ProtoTree, ProtoPFormer) provide case-based, class-tied explanations but are trained end-to-end for specific tasks \citep{chen2019protopnet,nauta2021prototree,xue2022protopformer}.
Activation maximization and “feature visualization” (DeepDream-style) optimize inputs for chosen units to aid intuition, but produce synthetic, non-natural images \citep{mahendran2016visualizing,nguyen2016dgn,olah2017feature}.
These approaches excel at explaining predictions or testing for known concepts (confirmation) but cannot be used to discover patterns without prior specification (\Cref{fig:related-work}).

\paragraph{Sparse Autoencoders \& Decomposition.}
Classical dictionary learning and sparse coding demonstrated that sparse, parts-based factors can emerge from natural data \citep{olshausen1996nature,aharon2006ksvd,lee1999nmf}.
Recent work applies sparse autoencoders (SAEs) to model activations, showing that sparse decompositions can partially ``unmix'' superposed features and yield monosemantic directions \citep{elhage2022superposition,bricken2023monosemanticity,gao2025scaling, joseph_steering_2025} in language and vision models.
SAEs address the superposition hypothesis \citep{elhage2022superposition} by assigning features to directions in activation space rather than individual neurons.
\citep{gao2025scaling} demonstrate that $k$-sparse autoencoders scale effectively and improve reconstruction--sparsity tradeoffs. 
Furthermore, concurrent work argues for using SAEs for discovering unknown concepts rather than acting on known ones \citep{peng2025unknown} and applies dictionary learning to microscopy foundation models for biological concept extraction \citep{donhauser2024towards}.
Beyond SAEs, concept factorization methods such as CRAFT, ACE, and related work use NMF-like decompositions to extract human-interpretable concepts and quantify their importance \citep{fel2023craft,fel2023holistic, ghorbani2019ace}.
Finally, recent analyses highlight stability and identifiability limits of SAE dictionaries across runs \citep{leask2025sparse}, motivating careful evaluation protocols.

\paragraph{Discovery.}
Closest in intent to our goal are (i) work advocating SAEs for uncovering \emph{unknown} concepts rather than acting on \emph{known} ones \citep{peng2025unknown}, (ii) dictionary-learning approaches that extract biological concepts from microscopy or epigenetics foundation models  \citep{donhauser2024towards, wang_using_nodate}, and (iii) automated concept discovery pipelines that discover concepts and then name or validate them with limited supervision \citep{rao2024discover}.
We differ by offering a comprehensive analysis of SAEs for visual feature discovery, validating both on a standard semantic benchmark and an ecology case study and demonstrating how to use heuristics to enable discovery in practice.
We also differ from unsupervised dense segmentation methods (LOST, TokenCut, STEGO, MaskCut) by returning global features with per-patch activations rather than per-image masks or clusters \citep{simeoni2021lost,wang2023tokencut,hamilton2022stego,wang2023maskcut}. 

\section{Methodology} \label{sec:methodology}

\paragraph{Foundation Model.} We use the publicly available DINOv3 ViT-L/16 checkpoint \citep{simeoni2025dinov3}, additionally using Perception Encoder ViT-L/14 336 \citep{bolya2025perception} in ablations (\Cref{sec:ablations}). 
Images are resized with an aspect-ratio-aware transform that enforces exactly \num{256} patches per image (a $h\times w$ patch grid with $hw=256$ chosen to minimize aspect-ratio distortion), instead of forcing square $256\times256$ inputs.
We then extract patch-token activations ([CLS] and [REG] discarded). 

\paragraph{Sparse Autoencoders (SAEs).} 
Our dictionary learner is a ReLU SAE \citep{bricken2023monosemanticity,templeton2024scaling}; given a $d$-dimensional activation vector $\mathbf{x} \in \mathbb{R}^d$ from an intermediate layer $l$ of a transformer, an SAE maps $\mathbf{x}$ to a sparse representation $f(\mathbf{x})$ (\Cref{eq:enc,eq:act}) and reconstructs the original input (\Cref{eq:dec}):
\begin{align}
    \mathbf{h} &= W_\text{enc} (\mathbf{x} - b_\text{dec}) + b_\text{enc} \label{eq:enc} \\
    f(\mathbf{x}) &= \text{ReLU}(\mathbf{h}) \label{eq:act} \\
    \mathbf{\hat{x}} &= W_\text{dec} f(\mathbf{x}) + b_\text{dec} \label{eq:dec}
\end{align}
where $W_\text{enc} \in \mathbb{R}^{n \times d}$, $b_\text{enc} \in \mathbb{R}^{n}$, $W_\text{dec} \in \mathbb{R}^{d \times n}$ and $b_\text{dec} \in \mathbb{R}^{d}$.
The training objective minimizes reconstruction error while encouraging sparsity:
\begin{equation}
    \mathcal{L}(\theta) = ||\mathbf{x} - \mathbf{\hat{x}}||_2^2 + \lambda \mathcal{S}(f(\mathbf{x}))
\end{equation}
where $\lambda$ controls the sparsity penalty and $\mathcal{S}$ measures sparsity (L$_1$ norm for training, L$_0$ for model selection).

We train \num{16384}-latent SAEs for 100M patches from a single ViT layer and sweep a logarithmic grid of sparsity coefficients $\lambda$ and learning rates.
We summarize each trained model by its reconstruction quality (normalized MSE) and sparsity (L$_0$).
Normalized MSE (NMSE) is MSE normalized by a mean baseline MSE:
\begin{equation}\label{eq:nmse}
\text{NMSE} = \frac{\sum_i || \mathbf{x_i} - \mathbf{\hat{x}_i} ||_2^2}{\sum_i || \mathbf{x_i} - \mathbf{\bar{x}} ||_2^2}
\end{equation}
where $\mathbf{\hat{x}_i}$ is the predicted reconstruction and $\mathbf{\bar{x}}$ is the mean of all $\mathbf{x_i}$.

\paragraph{SAE variants.}\label{sec:ablation-topk}
We also use TopK \citep{gao2025scaling} and Matryoshka SAEs \citep{bussmann2025matryoshka} and investigate their impact on concept rediscovery. 

\textbf{Matryoshka SAEs} \citep{bussmann2025matryoshka} address feature splitting \citep{bricken2023monosemanticity,chanin2024absorption,leask2025sparse} by training multiple nested dictionaries simultaneously.
Each training step samples random prefixes of SAE latents and reconstructs the input using each prefix.
We sample ten prefixes $\mathcal{M} = {m_1, m_2, \ldots, m_{10}}$ with $m_i \in \{1,...,  16384\}$; the sum of prefix reconstruction losses replaces MSE in the objective:
\begin{equation}\label{eq:matryoshka-objective}
\mathcal{L}(\theta) = \sum_{m \in \mathcal{M}} ||\mathbf{x}-\mathbf{\hat{x}}_{0:m}||^2_2 + \lambda \mathcal{S}(f(\mathbf{x}))
\end{equation}
where $\mathbf{\hat{x}}_{0:m}$ is the reconstruction using only the first $m$ latents.
This encourages early latents to capture more general features, implicitly ordering them by decreasing reconstruction importance (analogous to PCA).
At inference, prefixes are not sampled, making Matryoshka SAEs function identically to vanilla SAEs.

\textbf{TopK SAEs} \citep{gao2025scaling} replace the ReLU activation and sparsity objective $\mathcal{S}$ with a TopK selection over latents, retaining only the $k$ largest activations per input and zeroing the rest.
This enforces a fixed sparsity level and improves feature selectivity relative to ReLU at comparable reconstruction error.

\paragraph{Ranking Heuristic.} \label{sec:heuristic}
When applying SAEs for visual feature discovery in \textit{Heliconius} butterflies, we focus on finding diagnostic traits between two species.
For each SAE feature, we first reduce its activation map for a given image to a single scalar by taking the maximum activation value across all spatial locations (patches). 
We then score each SAE feature for species separation using a rank-based separability statistic built from the Mann-Whitney U-statistic \citep{mann_test_1947, mason_areas_2002}, without training a classifier. 
We call this metric \textit{selectivity}: a value of 1 denotes perfect separation between subspecies, and 0 denotes chance-level separation.
\begin{align}
    \text{sel.} = |\frac{U}{n_1n_2} - 0.5| * 2
\end{align}
where $U$ is the Mann-Whitney U-statistic calculated using the reduced SAE feature values and $n_1, n_2$ are the counts for each respective species. See \Cref{app:metric-defs} for a full definition.

\begin{figure*}[t]
    \small
    \centering
    \begin{subfigure}[b]{0.23\textwidth}
        \centering
        \includegraphics[width=0.48\textwidth]{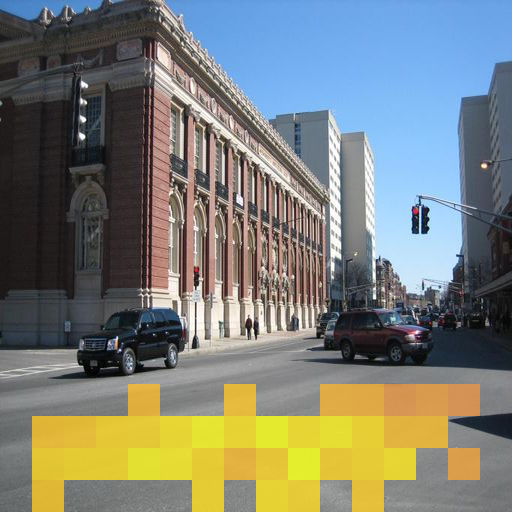}
        \hfill
        \includegraphics[width=0.48\textwidth]{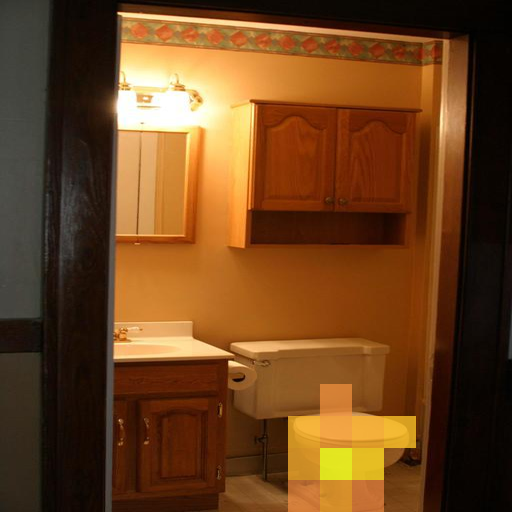}
        \includegraphics[width=0.48\textwidth]{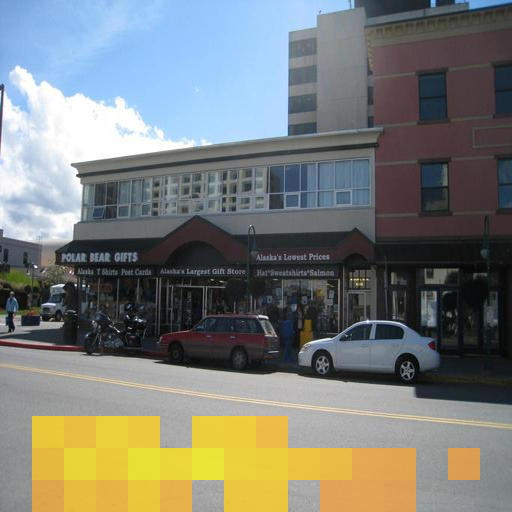}
        \hfill
        \includegraphics[width=0.48\textwidth]{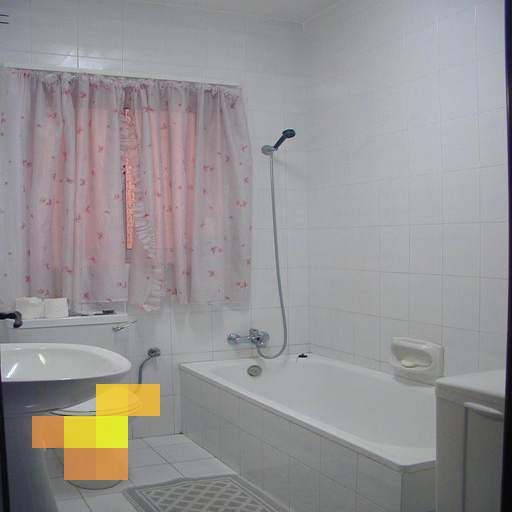}
        \caption{$k$-means}
        \label{fig:ade20k-examples-kmeans}
    \end{subfigure}
    \hfill
    \begin{subfigure}[b]{0.23\textwidth}
        \centering
        \includegraphics[width=0.48\textwidth]{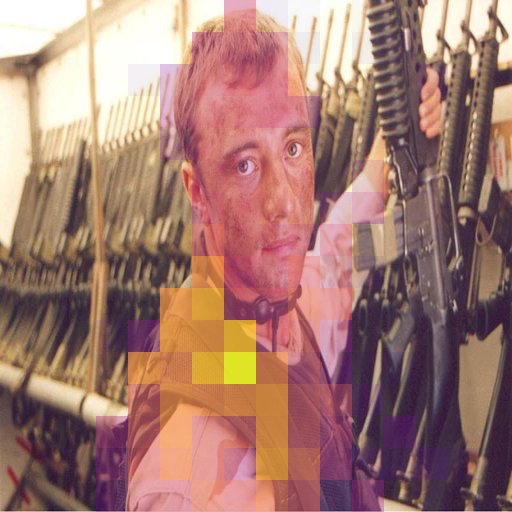}
        \hfill
        \includegraphics[width=0.48\textwidth]{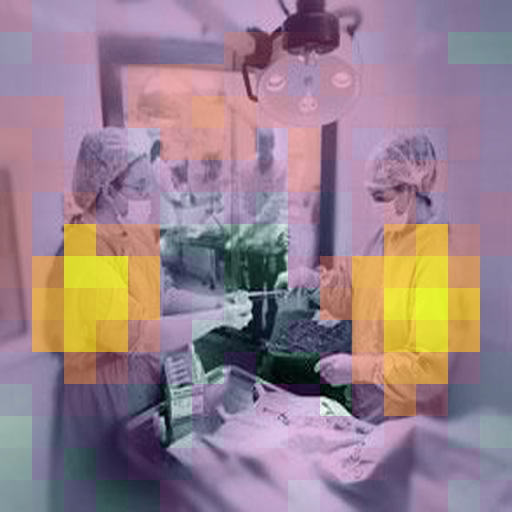}
        \includegraphics[width=0.48\textwidth]{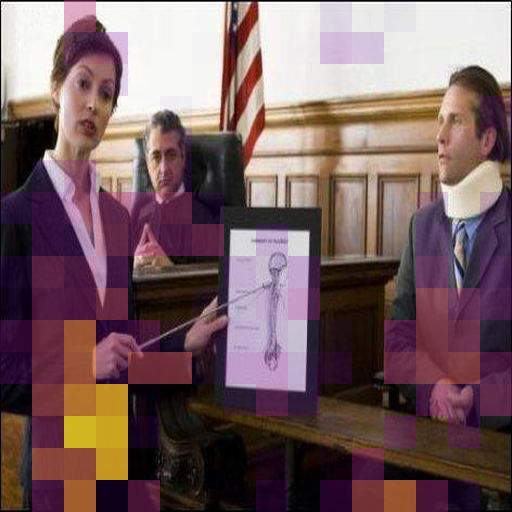}
        \hfill
        \includegraphics[width=0.48\textwidth]{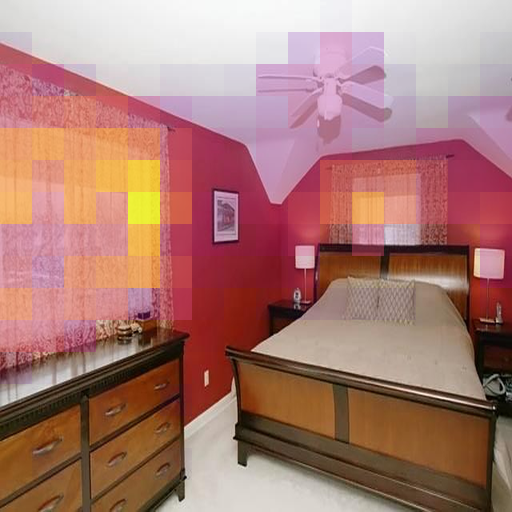}
        \caption{PCA}
        \label{fig:ade20k-examples-pca}
    \end{subfigure}
    \hfill
    \begin{subfigure}[b]{0.23\textwidth}
        \centering
        \includegraphics[width=0.48\textwidth]{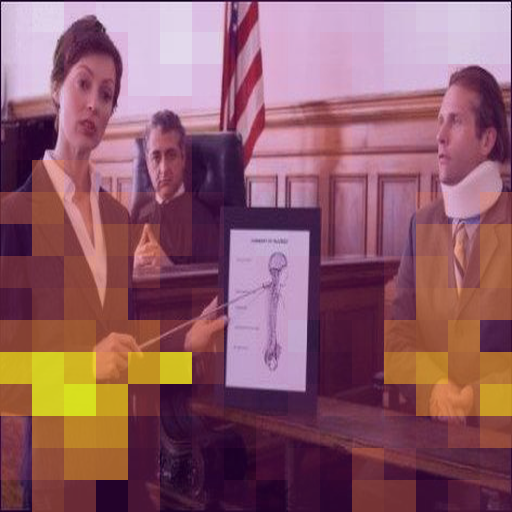}
        \hfill
        \includegraphics[width=0.48\textwidth]{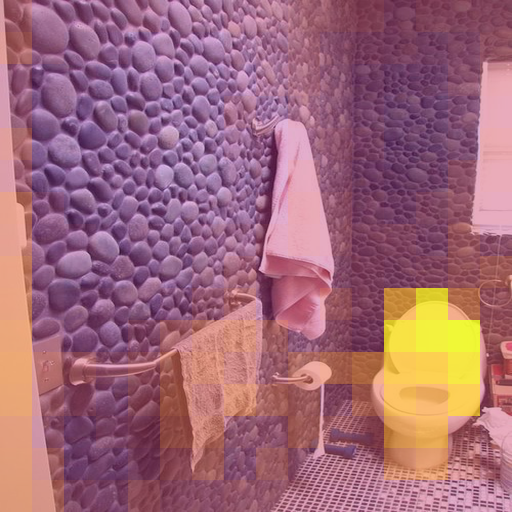}
        \includegraphics[width=0.48\textwidth]{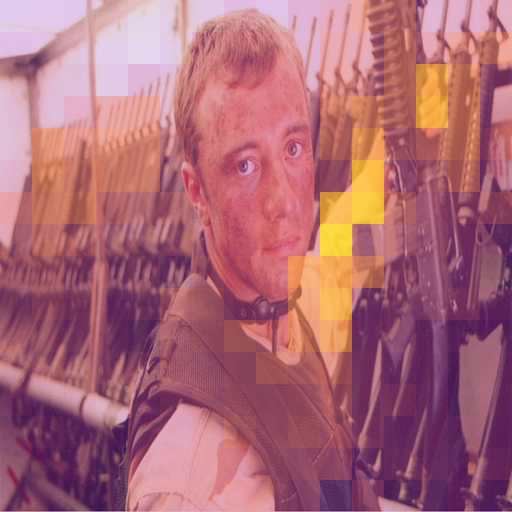}
        \hfill
        \includegraphics[width=0.48\textwidth]{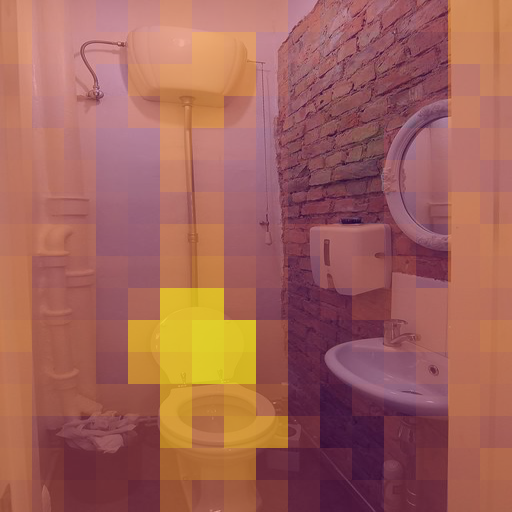}
        \caption{SemiNMF}
        \label{fig:ade20k-examples-seminmf}
    \end{subfigure}
    \hfill
    \begin{subfigure}[b]{0.23\textwidth}
        \centering
        \includegraphics[width=0.48\textwidth]{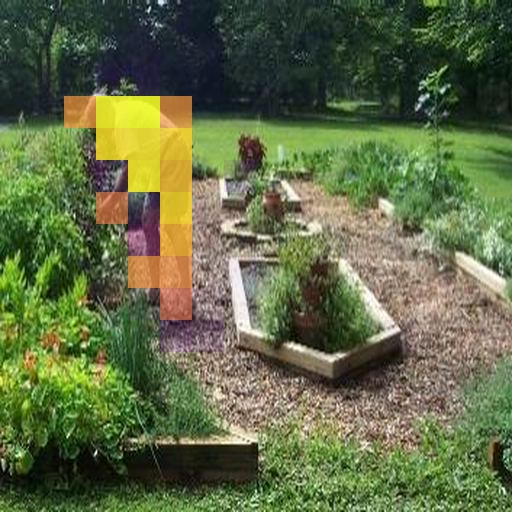}
        \hfill
        \includegraphics[width=0.48\textwidth]{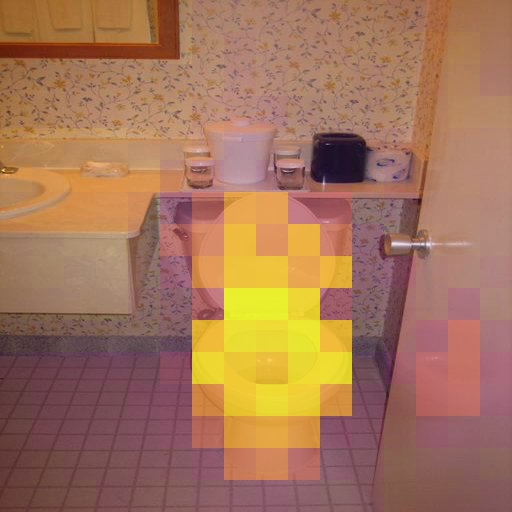}
        \includegraphics[width=0.48\textwidth]{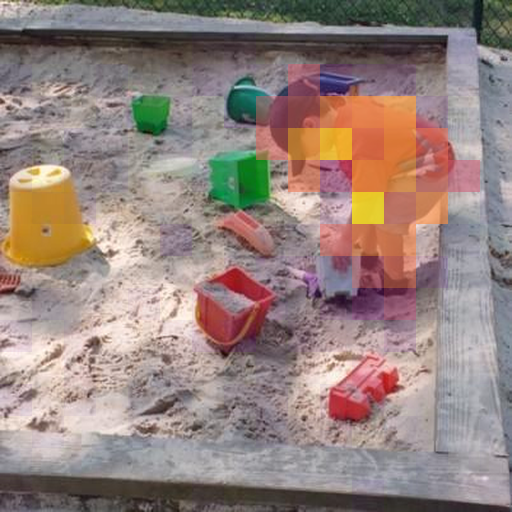}
        \hfill
        \includegraphics[width=0.48\textwidth]{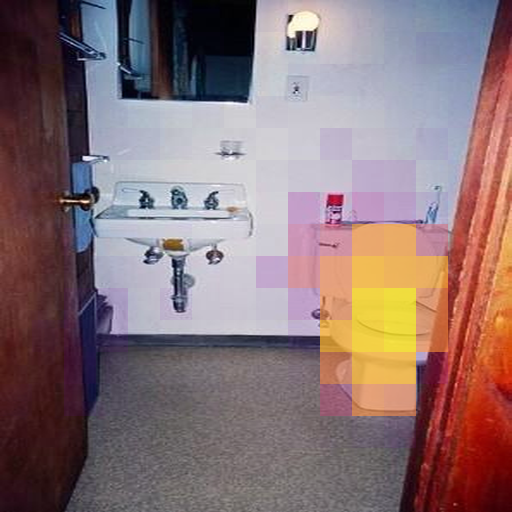}
        \caption{SAE}
        \label{fig:ade20k-examples-sae}
    \end{subfigure}
    \caption{   
    Top images for lowest-loss probes for ``person'' (left) and ``toilet'' (right) classes for each method.
    \textbf{(a):} $k$-means does not recover the ``person'' class; the best ``person'' cluster fires on roads.
    \textbf{(b):} PCA learns a ``person'' component, but does not consistently activate on the entire person. Furthermore, PCA does not recover the ``toilet'' class; the best ``toilet''  doesn't have an obvious semantic concept. 
    \textbf{(c):} Semi-nonnegative matrix factorization (SemiNMF) is a strong baseline and learns both ``person'' and ``toilet'' concepts.
    \textbf{(d):} SAEs reliably recover both ``person'' and ``toilet'' and activate on the entire object.
    \textbf{Takeaway}: While $k$-means and PCA report strong dictionary learning metrics, concepts recovered by SAEs are both more visually consistent and salient.
    }\label{fig:ade20k-examples}
\end{figure*}

\section{Experiments}\label{sec:experiments}

We evaluate the proposed method for open-ended visual discovery quality through \emph{controlled rediscovery} of ground-truth concepts withheld from SAE training.
Strong alignment on known concepts indicates that the same procedure is a promising method for surfacing meaningful structure where ground truth is unavailable.

Discovery is aided by monosemantic features: each latent should align with a single coherent concept.
We measure concept alignment with 1-D logistic probes, the standard evaluation in the SAE literature \citep{gao2025scaling}.
We report (1) probe $R$: loss $\mathcal{L}$ normalized by a bias-only baseline $\mathcal{L}_\pi$: $1 - (\mathcal{L}/\mathcal{L}_\pi)$, (2) mean Average Precision (mAP) across all classes, (3) Purity@$k$: precision of the top-$k$ activated patches for each feature, and (4) Coverage@$\tau$: the fraction of classes whose best-aligned feature achieves AP $\geq \tau$.
We select the SAE with the best probe loss on the training split and report validation metrics.
For exact definition of all metrics, see \Cref{app:metric-defs}.

\subsection{General semantic rediscovery}\label{sec:ade20k}

We use ADE20K \cite{zhou2017ade20k} as a controlled testbed for rediscovery: can label-free SAE features align with segmentation classes? 
SAEs are trained on ImageNet-1K activations only; ADE20K labels enter solely at evaluation, isolating extraction from supervision.
Concretely, we extract activations on ImageNet-1K (train/val) and ADE20K (train/val) from the final layer of DINOv3 ViT-L/16.

\paragraph{Baseline Methods.}
We compare SAEs to standard unsupervised baselines: $k$-means, principal component analysis (PCA), and semi-nonnegative matrix factorization \citep[SemiNMF;][]{ding2008convex}.
We fit all baselines on ImageNet-1K training activations.
For $k$-means, we use $k$=\num{16384} clusters and use the nearest cluster as the predicted reconstruction with a fixed L$_0$=$1$.
We fit PCA with $n$ components $n \in \{1,4,16,64,256,1024\}$.
For SemiNMF, we use a mini-batch semi-NMF estimator with nonnegative codes and an unconstrained dictionary; codes are inferred by multiplicative updates with cached positive/negative Gram terms, while the dictionary is updated from accumulated $Z^\top Z$ and $Z^\top A$ statistics with a ridge term.
For all methods, we report normalized reconstruction MSE on ImageNet-1K (validation split).
See \Cref{app:baselines} for more details.

\paragraph{Results.}
We compare all methods on the reconstruction--sparsity tradeoff in \Cref{fig:mse-l0} (right) and \Cref{fig:baseline-pareto-and-layerwise} (left), report rediscovery metrics in \Cref{tab:ade20k}, and show qualitative examples in \Cref{fig:ade20k-examples}.
SAEs do not improve upon normalized MSE and sparsity, but \textit{do} lead to better concept discovery as measured by mAP, Purity@$k$, and Coverage@$\tau$ when compared to other baseline methods.

\begin{table}[t]
    \centering
    \small
    \caption{SAEs vs.\ unsupervised baselines for rediscovering ADE20K's semantic segmentation classes from DINOv3 ViT-L/16.
    For each method, we choose the best layer from DINOv3 based on probe loss (\Cref{eq:probe-loss}) on ADE20K's training split and report ADE20K validation metrics.
    Probe $R$ is normalized probe loss, as described in \Cref{sec:experiments}.
    Purity@$k$ is precision over the top-$k$ highest-scoring patches for a feature, where $k=16$.
    Coverage@$\tau$ is the fraction of ADE20K classes with best AP $\geq\tau$, where $\tau= 0.3$.
    \textbf{Takeaway:} Even a vanilla ReLU SAE improves downstream concept rediscovery over traditional unsupervised methods, despite comparable reconstruction metrics. 
    }\label{tab:ade20k}
    \setlength{\tabcolsep}{3pt}
    \begin{tabular}{l rr rr rrrr}
    \toprule
    & \multicolumn{2}{c}{Source Recon.\ $\downarrow$} & \multicolumn{2}{c}{Target Recon.\ $\downarrow$} & \multicolumn{4}{c}{Downstream $\uparrow$} \\
    \cmidrule(lr){2-3} \cmidrule(lr){4-5} \cmidrule(lr){6-9}
    Method & NMSE & L$_0$ & NMSE & L$_0$ & Probe $R$ & mAP & Pur@$k$ & Cov@$\tau$ \\
    \midrule
    $k$-means & 0.508 & \textbf{1} & 0.618 & \textbf{1} & 0.001 & 0.031 & \textbf{0.795} & 0.026 \\  
    PCA & 0.288 & 256 & 0.352 & 256 & 0.259 & 0.093 & 0.612 & 0.079 \\ 
    SemiNMF & 0.298 & 256 & 0.363 & 256 & 0.209 & 0.088 & 0.696 & 0.079 \\
    \midrule
    SAE & \textbf{0.166} & 379.2 & \textbf{0.211} & 379.2 & \textbf{0.441} & \textbf{0.224} & 0.748 & \textbf{0.264} \\  
    \bottomrule
    \end{tabular}
\end{table}

\paragraph{Discussion.}
$k$-means and PCA are competitive on reconstruction--sparsity trade-offs, yet underperform SAEs on downstream concept-alignment metrics such as mAP, Purity@$k$, and Coverage@$\tau$. 
This reinforces that normalized MSE and L$_0$ measure how well activations are compressed, not how well individual latents track semantic concepts.
Prior work in evaluating SAEs for language model interpretability also finds dictionary-learning metrics insufficient \citep{karvonen2025saebench}.

%

\paragraph{Ablations.}\label{sec:ablations}
We ablate ViT size, ViT layer, ViT family and SAE variants to understand what drives concept rediscovery on ADE20K. 
We find that Matryoshka SAEs with TopK perform the best on downstream metrics when trained on ImageNet-1K patch activations and evaluated on ADE20K, achieving the highest probe $R$, mAP, and Coverage@$\tau$.
DINOv3 substantially outperforms Perception Encoder on all rediscovery metrics (\Cref{tab:ablations}), suggesting that self-supervised objectives like DINOv3 produce representations more amenable to sparse decomposition than contrastive vision-language training.
Additional ablations are found in \Cref{app:ablation-results}.

\begin{table}[t]
    \centering
    \small
    \setlength{\tabcolsep}{3pt}
    \caption{
    SAE variant and backbone ablation on ADE20K.
    Top rows compare SAE variants on DINOv3 ViT-L/16 (final layer); bottom row switches to a pretrained Perception Encoder ViT-L/14 checkpoint, using the Matryoshka+TopK SAE training protocol.
    Best SAE per configuration selected by training probe loss.
    }
    \label{tab:ablations}
    \begin{tabular}{l rr rr rrrr}
    \toprule
    & \multicolumn{2}{c}{Source Recon.\ $\downarrow$} & \multicolumn{2}{c}{Target Recon.\ $\downarrow$} & \multicolumn{4}{c}{Downstream $\uparrow$} \\
    \cmidrule(lr){2-3} \cmidrule(lr){4-5} \cmidrule(lr){6-9}
    Method & NMSE & L$_0$ & NMSE & L$_0$ & Probe $R$ & mAP & Pur@$k$ & Cov@$\tau$ \\
    \midrule
    ReLU & {0.166} & 379.2 & {0.211} & 379.2 & 0.441 & 0.224 & 0.748 & 0.264 \\  
    +Matryoshka & 0.591 & \textbf{6.5} & 0.702 & 6.5 & 0.402 & 0.319 & \textbf{0.808} & 0.405 \\  
    +TopK & 0.207 & 64 & 0.274 & 64 & \textbf{0.479} & \textbf{0.389} & 0.788 & \textbf{0.576} \\ 
    \midrule
    PE ViT-L/14 & \textbf{0.120} & 128 & \textbf{0.149} & 128 & 0.123 & 0.110 & 0.557 & 0.092 \\ 
    \bottomrule
    \end{tabular}
\end{table}

\subsection{Domain-specific semantic rediscovery}
\label{sec:fishvista}

\begin{table}[t]
    \centering
    \small
    \setlength{\tabcolsep}{3pt}
    \caption{Matryoshka TopK SAEs for concept rediscovery on general domain (ADE20K) and ecological imagery (FishVista) under the same DINOv3 ViT-L/16 backbone.
    Source Recon.\ measures reconstruction on the SAE's training distribution; Target Recon.\ measures transfer to the evaluation domain.}
    \label{tab:fishvista}
    \begin{tabular}{l rr rr rrrr}
    \toprule
    & \multicolumn{2}{c}{Source Recon.\ $\downarrow$} & \multicolumn{2}{c}{Target Recon.\ $\downarrow$} & \multicolumn{4}{c}{Downstream $\uparrow$} \\
    \cmidrule(lr){2-3} \cmidrule(lr){4-5} \cmidrule(lr){6-9}
    Dataset & NMSE & L$_0$ & NMSE & L$_0$ & Probe $R$ & mAP & Pur@$k$ & Cov@$\tau$ \\
    \midrule
    ADE20K & 0.207 & 64 & 0.274 & 64 & {0.479} & {0.389} & 0.788 & {0.576} \\  
    FishVista & 0.184 & 16 & 0.184 & 16 & 0.438 & 0.543 & 0.501 & 0.7 \\ 
    \bottomrule
    \end{tabular}
\end{table}

\begin{figure*}[t]
    \small
    \centering
    \begin{subfigure}[b]{0.32\textwidth}
        \centering
        \includegraphics[width=0.48\textwidth]{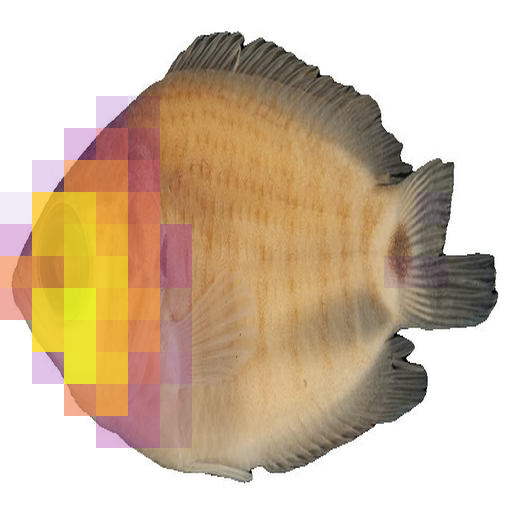}
        \hfill
        \includegraphics[width=0.48\textwidth]{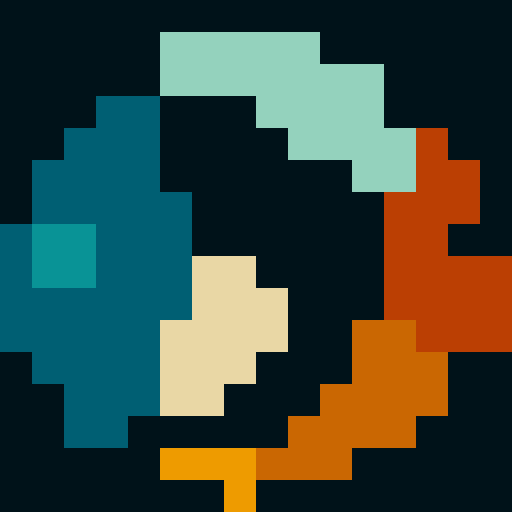}
        \caption{\textbf{\textcolor{blue}{``Head'' (\texttt{DINOv3-16K/9})}}}
        \label{fig:fishvista-head}
    \end{subfigure}
    \hfill
    \begin{subfigure}[b]{0.34\textwidth}
        \centering
        \includegraphics[width=0.48\textwidth]{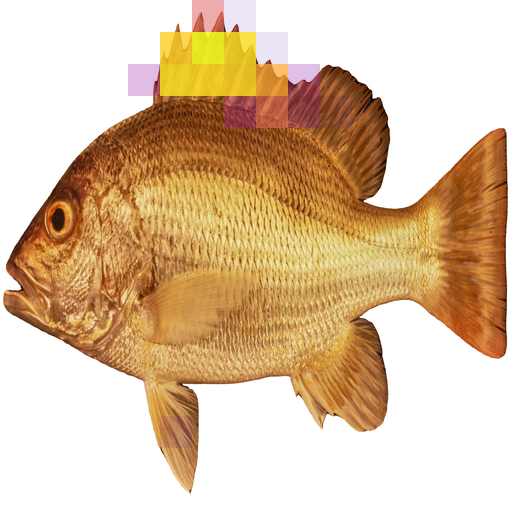}
        \hfill
        \includegraphics[width=0.48\textwidth]{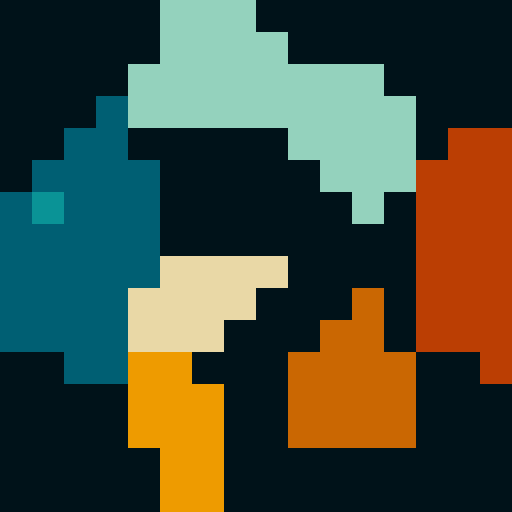}
        \caption{\textbf{\textcolor{sea}{``Dorsal fin'' (\texttt{DINOv3-16K/13})}}}
        \label{fig:fishvista-dorsal}
    \end{subfigure}
    \hfill
    \begin{subfigure}[b]{0.32\textwidth}
        \centering
        \includegraphics[width=0.48\textwidth]{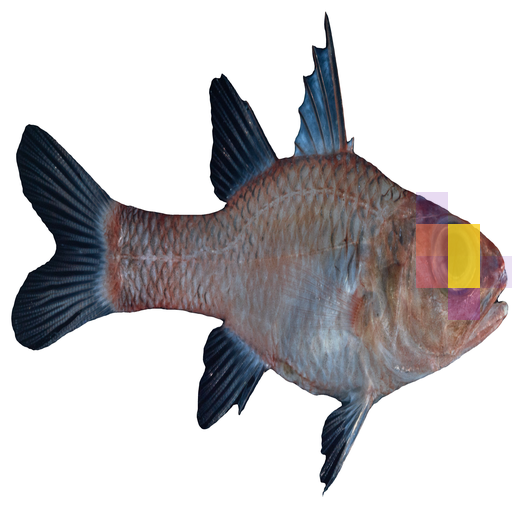}
        \hfill
        \includegraphics[width=0.48\textwidth]{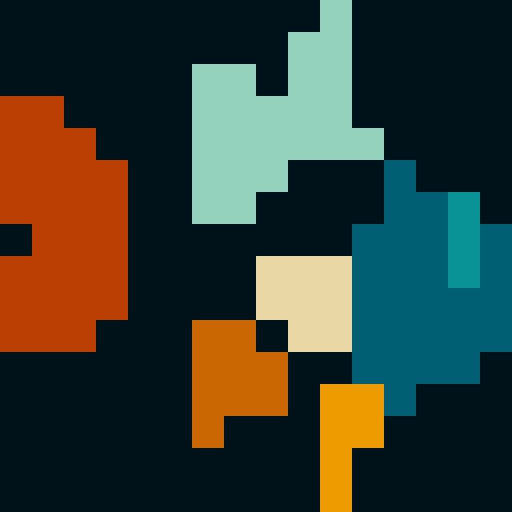}
        \caption{\textbf{\textcolor{cyan}{``Eye'' (\texttt{DINOv3-16K/69})}}}
        \label{fig:fishvista-eye}
    \end{subfigure}
    \caption{
    Example TopK SAE features (left) and patch-level segmentation masks (right) for \textbf{\textcolor{blue}{(a) ``Head''}}, \textbf{\textcolor{sea}{(b) ``Dorsal fin''}} and \textbf{\textcolor{cyan}{(c) ``Eye''}} from DINOv3 ViT-L/16.
    Features were picked by minimizing cross entropy on binary classification for each body part, as described in \Cref{sec:fishvista,eq:probe-loss}.
    \textbf{Takeaway}: TopK SAEs accurately learn scientific concepts (anatomical body parts) without labels. 
    }\label{fig:fishvista-examples}
\end{figure*}

To validate this approach beyond general-domain semantic concepts, we test it in ecological imagery: can vision foundation models, in combination with SAEs, rediscover the anatomy of diverse fish species? 
We use FishVista \citep{mehrab2024fishvista,fishvistadataset} and the 10 annotated body-part traits to answer this question (see \Cref{fig:fishvista-examples} for dataset examples).
We use the final DINOv3+TopK setting from \Cref{sec:ablations}, training SAEs on \num{56.3}K fish images and evaluating SAE discovery on the \num{4.6}K (disjoint) images with body part segmentations.

\paragraph{Results.}

\begin{figure*}[t]
    \small
    \centering
    \begin{subfigure}[b]{0.48\textwidth}
        \includegraphics[width=\linewidth]{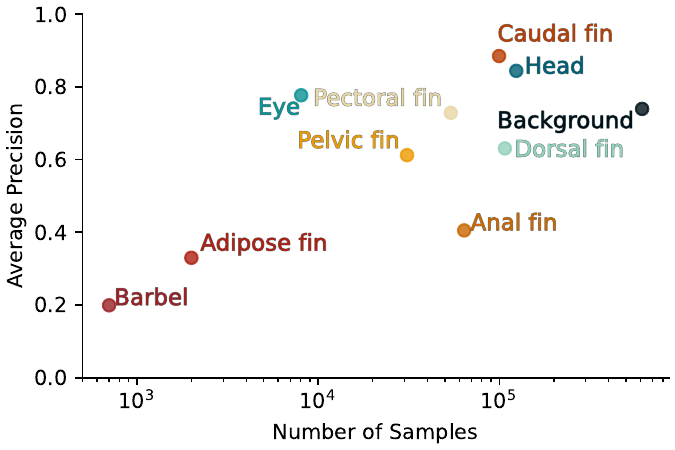}
    \end{subfigure}
    \hfill
    \begin{subfigure}[b]{0.48\textwidth}
        \centering
        \includegraphics[width=\linewidth]{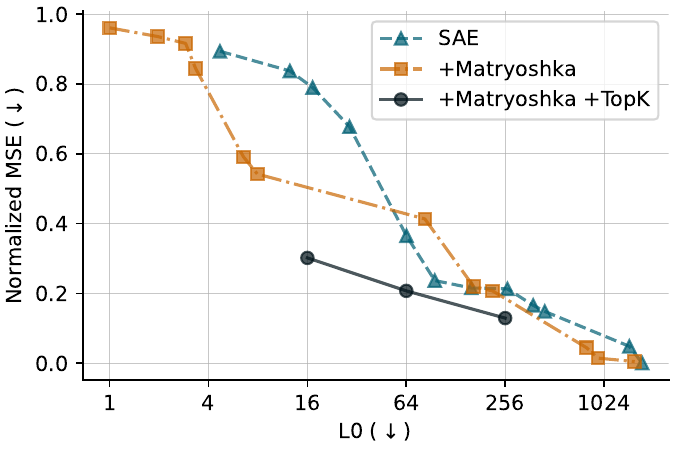}
    \end{subfigure}
    \caption{\textbf{(Left)} Prevalence vs rediscovery on FishVista. 
    Each point is a body part; x-axis: number of patch-level samples in the validation split, y-axis: AP of the best-matching (lowest probe loss) TopK SAE latent.
    More common body parts are rediscovered more reliably. 
    \textbf{Takeaway:} Higher concept prevalence is associated with more rediscovery.
    \textbf{(Right)} Reconstruction--sparsity tradeoff for SAE variants (ReLU, Matryoshka, TopK) on DINOv3 ViT-L/16 ImageNet-1K activations. TopK achieves the best reconstruction at fixed sparsity.
    }\label{fig:prevalence}\label{fig:mse-l0}
\end{figure*}

We find that DINOv3+Matryoshka TopK SAEs reliably find domain-specific concepts; we compare results on ADE20K and FishVista in \Cref{tab:fishvista}, show examples in \Cref{fig:fishvista-examples} and compare concept prevalence with average precision in \Cref{fig:prevalence} (left).
These results indicate that strong foundation models, in combination with SAEs, can surface features that align closely with expert-defined anatomical annotations.

\paragraph{Discussion.}
Body parts that occur in more patches are recovered with higher average precision (\Cref{fig:prevalence}, left), while rare parts are substantially harder to rediscover. 
This implies that scientific applications targeting rare or long-tail structures and patterns may need either more data, targeted sampling, or novel dictionary learning algorithms that learn rare structures more efficiently.

\subsection{Question-driven feature ranking} \label{sec:butterfly-results}
Previously, we established that SAEs are a strong decomposition method for learning visual features.
\textit{Heliconius} butterflies are an often-studied example of mimicry and evolution \citep{mcmillan_patterning_2020}; understanding subspecies wing patterns and traits is of great interest to researchers \citep{hines_wing_2011, jiggins_ecological_2008, mcmillan_patterning_2020}. 
Natural selection has driven striking convergence in the appearance of different \textit{Heliconius} butterfly species, such that different species can appear almost identical \citep{reed_optix_2011}. 
Nonetheless, decades of work by biologists have revealed certain key traits, such as subtle differences in color pattern, that can be used to distinguish mimics \citep{muller1879ituna, devries_butterflies_1987, jiggins_ecological_2008, lawrence_perceptual_2026}. 
We propose using SAEs with our \textit{selectivity} heuristic (\Cref{sec:heuristic}) to identify visual features of \textit{Heliconius} wings that may separate mimic subspecies.
To test the efficacy of using this approach in practice, we test whether it can efficiently rediscover known diagnostic traits between \textit{Heliconius erato} and \textit{Heliconius melpomene} subspecies.
We train Matryoshka TopK SAEs on DINOv3 layer 22 and 24 patch activations of 24.8K \textit{Heliconius} images \citep{hdr_imageomics_institute_heliconius-collection_cambridge-butterfly_nodate} and rank features for selectivity between three mimic pairs.
We then compare the highest selectivity score across all values of $k$ in the Pareto-optimal SAEs.

\begin{figure*}[t]
    \small
    \centering
    \begin{subfigure}[b]{0.32\textwidth}
        \centering
        \includegraphics[width=0.98\textwidth]{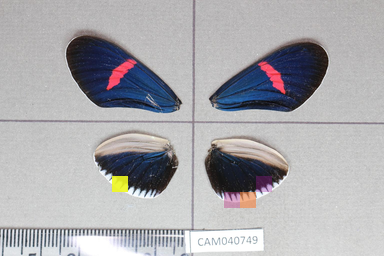}
        \includegraphics[width=0.98\textwidth]{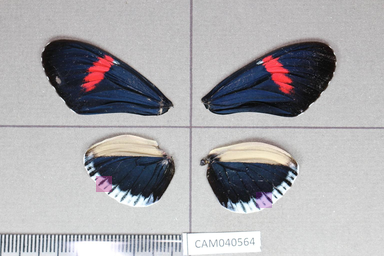}
        \caption{\textbf{\textcolor{blue}{\texttt{Feature 1,662}}} (sel.=0.96)}
        \label{fig:butterfly-cy}
    \end{subfigure}
    \hfill
    \begin{subfigure}[b]{0.32\textwidth}
        \centering
        \includegraphics[width=0.98\textwidth]{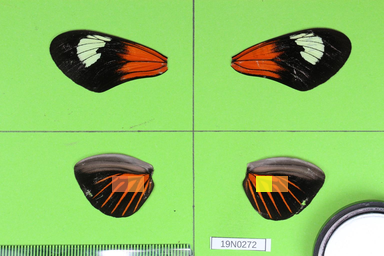}
        \includegraphics[width=0.98\textwidth]{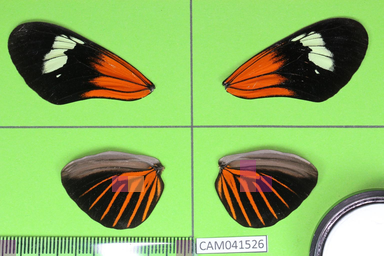}
        \caption{\textbf{\textcolor{sea}{\texttt{Feature 104}}} (sel.=0.99)}
        \label{fig:butterfly-lat}
    \end{subfigure}
    \hfill
    \begin{subfigure}[b]{0.32\textwidth}
        \centering
        \includegraphics[width=0.98\textwidth]{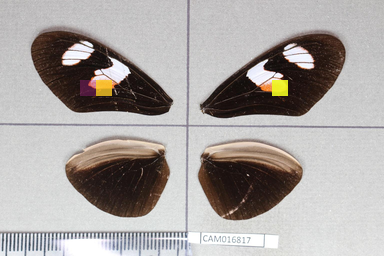}
        \includegraphics[width=0.98\textwidth]{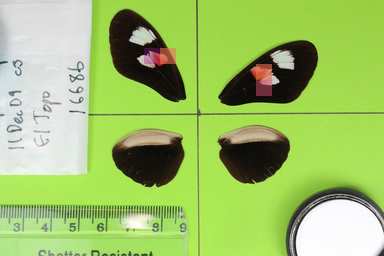}
        \caption{\textbf{\textcolor{cyan}{\texttt{Feature 192}}} (sel.=0.94)}
        \label{fig:butterfly-not}
    \end{subfigure}
    \caption{
    For a single Matryoshka TopK SAE trained on DINOv3 ViT-L/16 patch activations, the highest selectivity features for each studied \textit{H. erato} (top)/\textit{H. melpomene} (bottom) mimic pair. \textbf{\textcolor{blue}{(a) Feature 1,662}} corresponds to hindwing fringe, \textbf{\textcolor{sea}{(b) Feature 104}} corresponds to hindwing ray shape, and \textbf{\textcolor{cyan}{(c) Feature 192}} corresponds to the forewing red patch placement. Note that the highest-ranked \textit{H. melpomene} image for \textbf{\textcolor{sea}{(b) Feature 104}} is actually a mislabeled \textit{H. erato}. Similar features were ranked very highly for all 3 pairs among other trained SAEs.
    \textbf{Takeaway}: SAE features can be easily ranked using relevant heuristics to identify the most promising discovery candidates.
    }\label{fig:butterfly-features}
\end{figure*}

First, we find that layer 22 consistently provides higher selectivity scores, indicating that the last layer may not be optimal for all cases.
Then, a co-author with domain expertise in \textit{Heliconius} wing-patterns manually annotated the 5 highest-ranked features from layer 22 pareto-optimal SAEs for each mimic pair, determining whether each corresponded to a known diagnostic trait.
We find that in all 3 mimic pairs, the Matryoshka TopK SAE features with the highest selectivity values correspond to known diagnostic traits  (\Cref{fig:butterfly-features}, \Cref{app:butterfly-images}) \citep{cast_diversite_2025, lawrence_perceptual_2026}.
Additionally, for each mimic pair, 60-100\% of the five highest-ranked features by selectivity corresponded to known diagnostic traits.
We provide images and annotations of the five highest-ranked features by selectivity for each mimic pair in \Cref{app:butterfly-images}.
Exact SAE training parameters are provided in \Cref{app:butterfly-saes}.

\paragraph{Discussion.}
The consistently high rank of features corresponding with expert-validated diagnostic traits in \textit{Heliconius} mimic pairs suggests that using well-designed heuristics enables quick identification of scientifically interesting features, without the need for fine-grained concept labels.
Additionally, any heuristic could easily be used in place of selectivity, allowing for researchers to select for any type of visual feature that may be interesting for a specific problem.

\section{Conclusion}

In the settings we studied, we demonstrated that sparse autoencoders can systematically recover semantic structure from vision foundation model representations without concept supervision, using controlled rediscovery on both general-domain visual concepts and a scientific case study.
On ADE20K, SAEs consistently rediscovered semantic units from DINOv3 activations trained only on ImageNet-1K, outperforming decomposition baselines ($k$-means, PCA \& SemiNMF) on almost all concept alignment metrics. 
On ecological images, the same procedure surfaced anatomical structures without using segmentation or part labels during extraction.
Finally, when used with simple ranking heuristics, features corresponding with known diagnostic traits were easily identified.
Our results provide evidence that SAEs are a practical method for extracting interpretable features from vision foundation models without needing supervision.
With this, the quantitative nature of SAE features shifts the search for interesting visual phenomena to a heuristic design problem, allowing researchers to better leverage foundation model representations and more quickly create meaningful hypotheses.
Importantly, our contributions should be understood as validating SAEs as an \textit{instrument} for open-ended visual discovery, not as delivering novel biological or physical discoveries themselves.
In the future, we aim to use this approach in collaboration with domain experts to extract truly novel scientific insights from vision foundation models.

\newpage

\section{AI Statement}\label{app:ai-statement}
In this work, we used generative AI tools to assist with editing software code, editing drafts, and reviewing existing literature. We have reviewed all AI-assisted work, including reviewing AI-generated code and manually reading any sources provided by AI when used for reviewing literature. We take responsibility for the final content of this work, including text, claims or artifacts produced with the aid of generative AI.

\section{Reproducibility Statement}
In order to ensure reproducibility of this work, all training and implementation details of SAEs, metrics, and ranking heuristics are provided in \Cref{sec:methodology}, \Cref{sec:experiments}, \Cref{app:hyperparameters}, \Cref{app:metric-defs}, and \Cref{app:butterfly-saes}. Additionally, code used for all experiments is provided at \url{https://github.com/Imageomics/saev}. 



\bibliographystyle{iclr2027_conference}
\bibliography{main}

\clearpage
\appendix

\setcounter{table}{0}
\renewcommand\thetable{\Alph{section}\arabic{table}}
\setcounter{figure}{0}
\renewcommand\thefigure{\Alph{section}\arabic{figure}}

\section*{Appendices}

\begin{enumerate}[nosep]
    \item{\Cref{app:limitations}: Limitations}
    \item{\Cref{app:hyperparameters}: Training details and hyperparameters}
    \item{\Cref{app:metric-defs}: Precise metric definitions}
    \item{\Cref{app:qualitative-examples}: Additional qualitative examples}
    \item{\Cref{app:butterfly-saes}: SAE training details for \textit{Heliconius} images}
    \item{\Cref{app:ablation-results}: Detailed ablation results}
    \item{\Cref{app:baselines}: Baseline methods}
    \item{\Cref{app:layer-metrics}: Additional layer-depth ablations}
    \item{\Cref{app:butterfly-images}: Additional butterfly feature images}
\end{enumerate}

\section{Limitations}\label{app:limitations}
While the proposed methodology effectively learns and identifies meaningful, relevant visual features for discovery, advances are still needed to better enable discovery.
First, our evaluation relies on ground-truth annotations to measure concept alignment, which limits our ability to assess discovery in truly unannotated domains. 
Second, previous work has demonstrated issues with SAE features, such as instability \citep{paulo_sparse_2025, papadimitriou_interpreting_2025, leask2025sparse} and splitting or absorption \citep{chanin2024absorption}.
With this, vision features can be difficult to interpret, and SAE features may not perfectly align with a human interpretation of them.
As such, it may be necessary to include additional mechanisms for interpretability \citep{shaham_multimodal_2024, he_saeexplainer_2026} or leverage multimodal models.
This approach also only provides \textit{possibly} interesting features, and does not perform fully-automatic scientific discovery. 
However, we argue that visual scientific discovery necessarily requires expert validation and reasoning.
Finally, while strong performance on two domains suggests the method will generalize, we leave both systematic evaluation of novel discoveries and applying SAEs to non-vision foundation models to future work.

\section{Training Details}\label{app:hyperparameters}

Our code contains instructions to reproduce our exact findings.

\paragraph{SAE Training.}
\begin{itemize}
    \item{Initialize $W_\text{enc}$ and $W_\text{dec}$ with Kaiming initialization \citep{kaiming2015init} and $b_\text{enc}$ and $b_\text{dec}$ with all zeros.}
    \item{Train SAEs with Adam \citep{kingma2014adam} for \num{100}M examples with a batch size of \num{16384}.}
    \item{Linearly warmup learning rate from \num{0} to the maximum over 500 steps, then use cosine decay to \num{0}.}
    \item{Linearly warmup the sparsity coefficient $\lambda$ from 0 to the maximum over all of training.}
    \item{Normalize $W_\text{dec}$ to have unit norm columns at every step and remove gradients parallel to the columns to avoid interactions between Adam moments and normalization \citep{bricken2023monosemanticity}.}
    \item{Sweep both learning rate and sparsity coefficient $\lambda$ for each setting, using the values in \Cref{tab:hyperparams}.}
\end{itemize}

\begin{table}[b]
    \small
    \centering
    \begin{tabular}{ll}
    \toprule
    Hyperparam & Sweep Values \\
    \midrule
    Learning Rate & [\num{3e-4}, \num{1e-3}, \num{3e-3}, \num{1e-2}, \num{3e-2}, \num{1e-1}]\\
    Sparsity Coeff. & [\num{1e-4}, \num{1e-3}, \num{1e-2}, \num{1e-1}] \\
    \bottomrule
    \end{tabular}
    \caption{Sparse autoencoder hyperparameter sweep values.}
    \label{tab:hyperparams}
\end{table}

\paragraph{Baseline Training.}
We train $k$-means and PCA with randomly shuffled mini-batches with a batch size of \num{16384} using \num{100}M examples.

\paragraph{1D Probe Training.}
We optimize \Cref{eq:probe-loss} for 1-dimensional logistic regression classifiers for each pair of \{SAE latent $i$, class $c$\} with a ridge of \num{1e-8} and \num{30} steps of Newton-style optimization.
We initialize $w$ to \num{0} and $b$ to the prevalence of the class $c$ in the entire dataset.

\section{Metric Definitions}\label{app:metric-defs}

\begin{figure*}[t]
    \small
    \centering
    \begin{subfigure}[b]{0.49\textwidth}
        \centering
        \includegraphics[width=\textwidth]{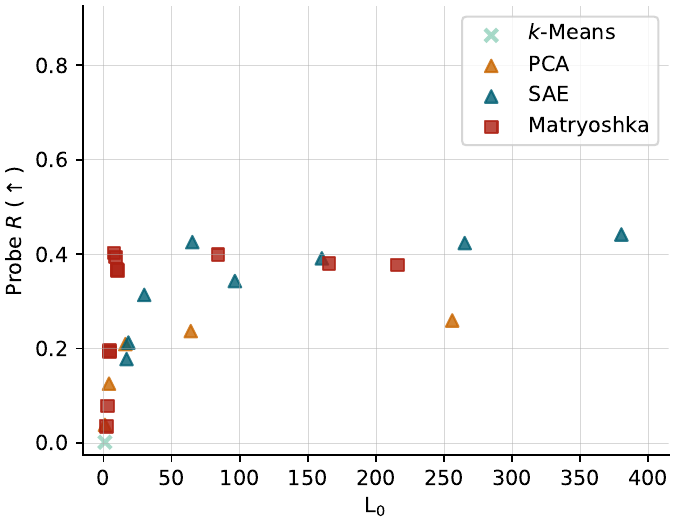}
        \caption{Probe $R$}
        \label{fig:ade20k-methods-r}
    \end{subfigure}
    \hfill
    \begin{subfigure}[b]{0.49\textwidth}
        \centering
        \includegraphics[width=\textwidth]{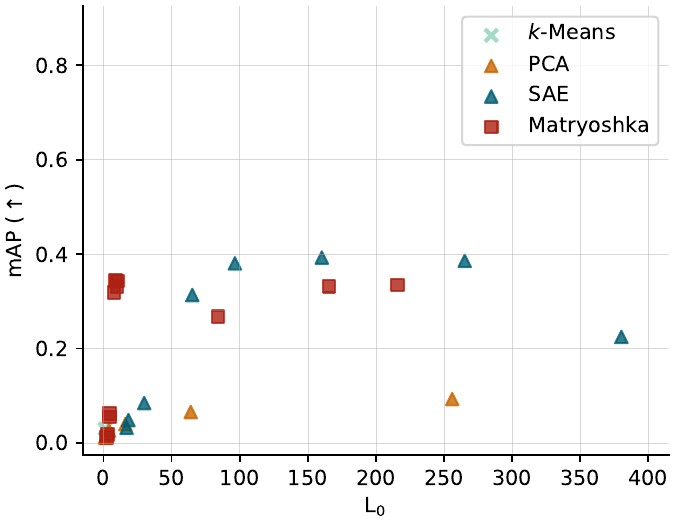}
        \caption{mAP}
        \label{fig:ade20k-methods-map}
    \end{subfigure}
    \begin{subfigure}[b]{0.49\textwidth}
        \centering
        \includegraphics[width=\textwidth]{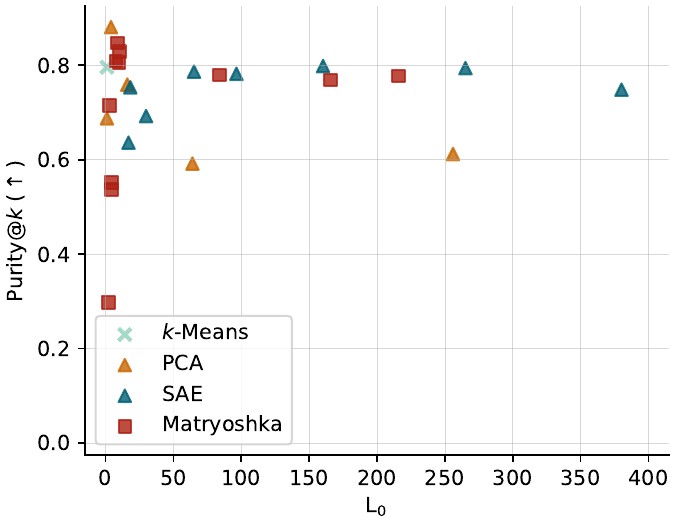}
        \caption{Purity@$k$}
        \label{fig:ade20k-methods-purity}
    \end{subfigure}
    \hfill
    \begin{subfigure}[b]{0.49\textwidth}
        \centering
        \includegraphics[width=\textwidth]{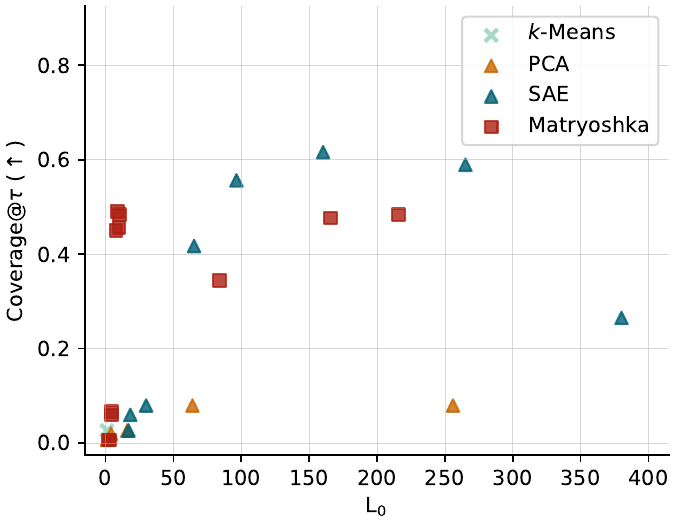}
        \caption{Coverage@$\tau$}
        \label{fig:ade20k-methods-cov}
    \end{subfigure}
    \caption{
    We compare $k$-means, PCA, vanilla SAEs and Matryoshka SAEs against sparsity (L$_0$) and downstream metrics (Probe $R$, mAP, Purity@$k$ and Coverage@$\tau$) for ADE20K.
    }
    \label{fig:ade20k-methods-metrics}
\end{figure*}

We propose \textit{selectivity} as a heuristic for identifying features corresponding to known diagnostic traits in \textit{Heliconius} butterflies. 
We use a variety of metrics to evaluate SAE probe quality.
We include scatter plots of our different metrics (Probe $R$, mAP, Purity@$k$ and Coverage@$\tau$) in \Cref{fig:ade20k-methods-metrics}.

\paragraph{Selectivity}
Let $x_1, ..., x_{n_1}$ and $y_1, ..., y_{n_2}$ be SAE feature values (max per image) corresponding to 2 classes.
Then, let $T$ be the sum of the ranks of values $y_1, ..., y_{n_2}$ in the ordered sequence of $x$'s and $y$'s. 
We then have:

\begin{align}
    U = n_1n_2 + \frac{n_2(n_2 + 1)}{2} -T
\end{align}

where $U$ is the Mann-Whitney $U$-statistic \citep{mann_test_1947}.
Therefore, in full we define \textit{selectivity} as:

\begin{align}
    \text{sel.} = |\frac{n_1n_2 + \frac{n_2(n_2 + 1)}{2} - T}{n_1n_2} - 0.5| * 2
\end{align}

\paragraph{Probe $R$.}
For each ground truth concept, we fit a binary logistic on each latent $i$'s activation $z_i$ and select the best latent per class:
\begin{equation}\label{eq:probe-loss}
    \min_{i,w,b} \mathbb{E} \big[-y \log \sigma (wz_i + b) - (1-y) \log (1 - \sigma (wz_i+b)) \big]
\end{equation}
Because the probe is axis-aligned by design, it favors representations whose individual axes carry semantic meaning.
This is not a limitation of the evaluation but a requirement of the use case: features that cannot be understood one at a time cannot be browsed for discovery.

We record the binary cross entropy loss  (see \Cref{eq:probe-loss}) for the trained probe ($\mathcal{L}$) and the loss for a bias-only probe ($\mathcal{L}_\pi$). 
We normalize the trained loss by the bias-only loss: $1 - (\mathcal{L}/\mathcal{L}_\pi)$.

\paragraph{mAP.}
For each class, we treat its patches as positives and all others as negatives, rank all validation patches by the corresponding probe score, and compute average precision as the area under the precision–recall curve.
We report mean average precision (mAP) as the macro-average of per-class AP values, assigning AP=\num{0} to classes without positive examples.

\paragraph{Purity@$k$.}
For each SAE latent $i$, we take the $k$ image patches that maximally activate latent $i$ and measure how many of the $k$ patches have the same patch label, then normalize by $k$ to get a score between $0$ and $1$.
We set $k$=$16$ in our work.

\paragraph{Coverage@$\tau$.}
We measure the fraction of classes (\num{151} for ADE20K, \num{10} for FishVista) that have an SAE latent with a minimum average precision of $\tau$.
We set $\tau$=\num{0.3} in our work.

\section{Qualitative Examples}\label{app:qualitative-examples}

\begin{figure*}[t]
    \small
    \centering
    \begin{subfigure}[b]{0.23\textwidth}
        \centering
        \includegraphics[width=0.48\textwidth]{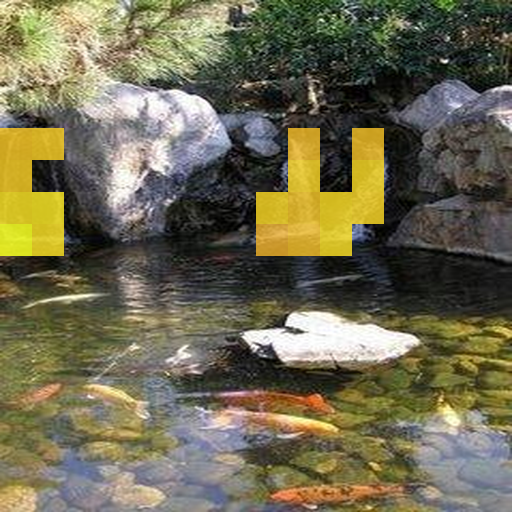}
        \hfill
        \includegraphics[width=0.48\textwidth]{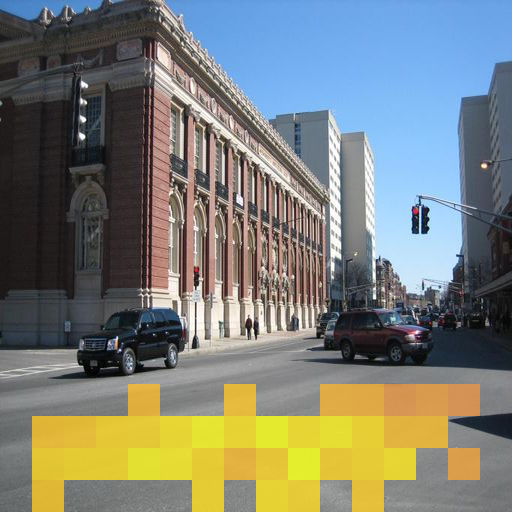}
        \includegraphics[width=0.48\textwidth]{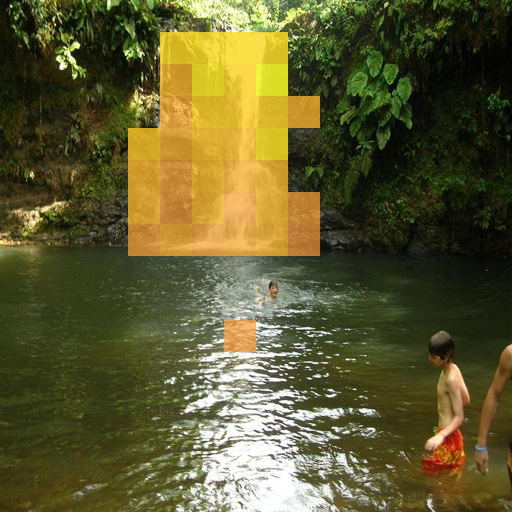}
        \hfill
        \includegraphics[width=0.48\textwidth]{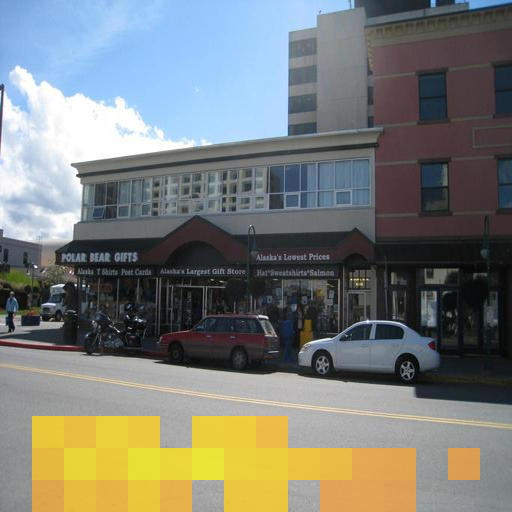}
        \includegraphics[width=0.48\textwidth]{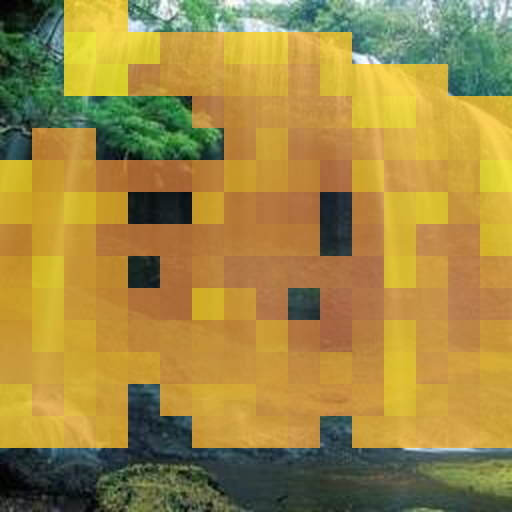}
        \hfill
        \includegraphics[width=0.48\textwidth]{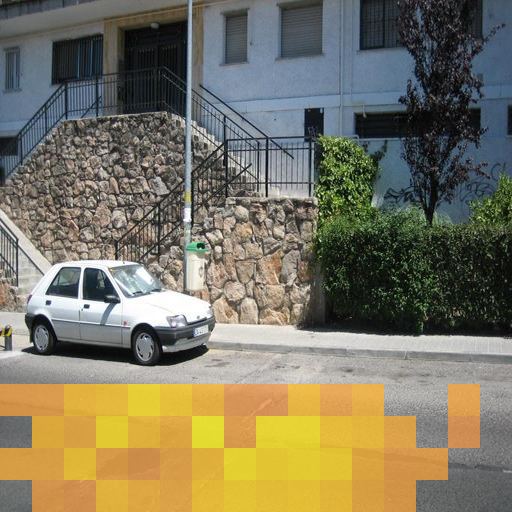}
        \includegraphics[width=0.48\textwidth]{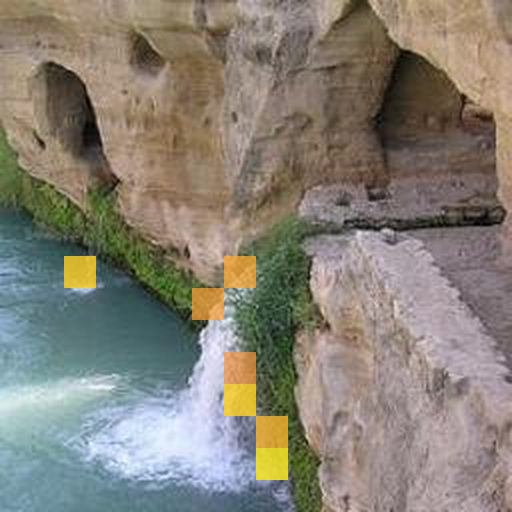}
        \hfill
        \includegraphics[width=0.48\textwidth]{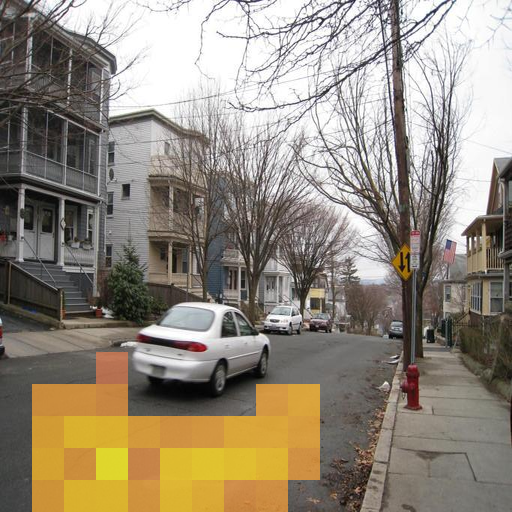}
        \caption{$k$-means}
        \label{fig:additional-ade20k-examples-kmeans}
    \end{subfigure}
    \hfill
    \begin{subfigure}[b]{0.23\textwidth}
        \centering
        \includegraphics[width=0.48\textwidth]{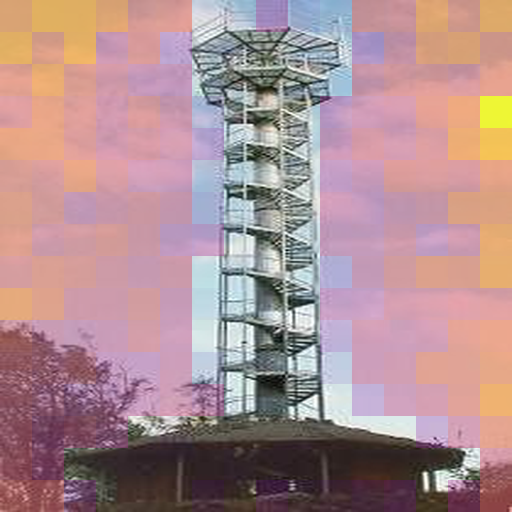}
        \hfill
        \includegraphics[width=0.48\textwidth]{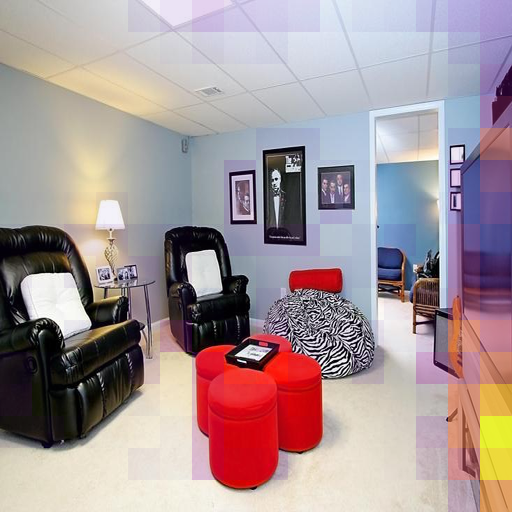}
        \includegraphics[width=0.48\textwidth]{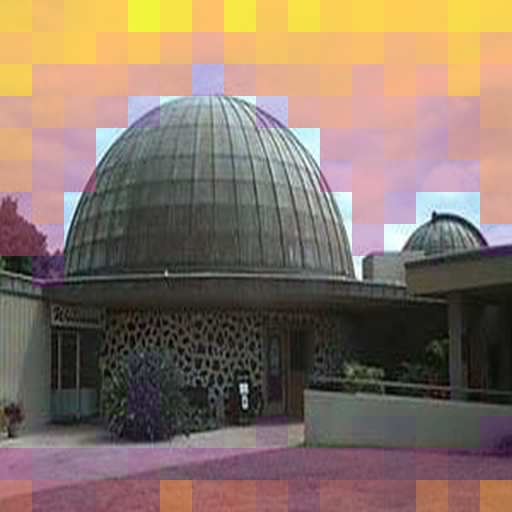}
        \hfill
        \includegraphics[width=0.48\textwidth]{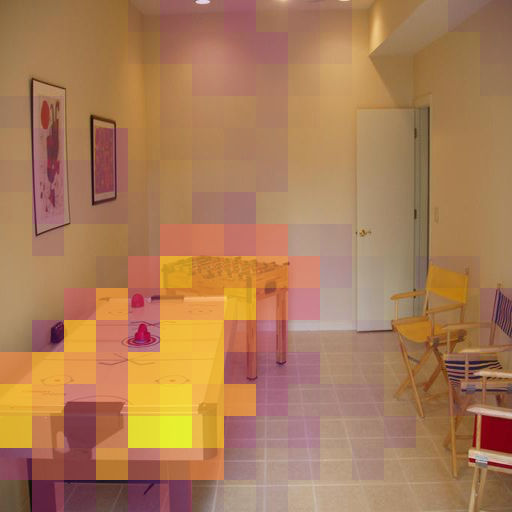}
        \includegraphics[width=0.48\textwidth]{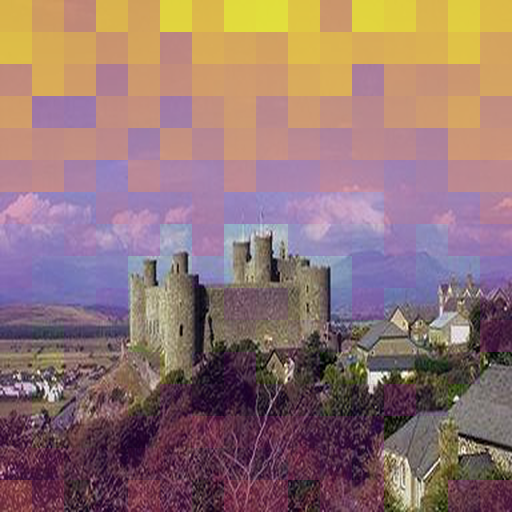}
        \hfill
        \includegraphics[width=0.48\textwidth]{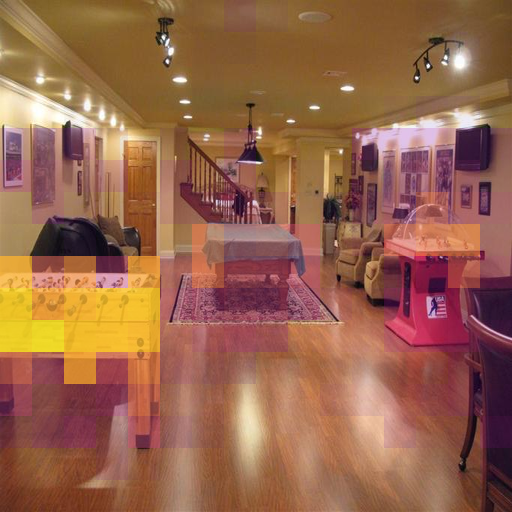}
        \includegraphics[width=0.48\textwidth]{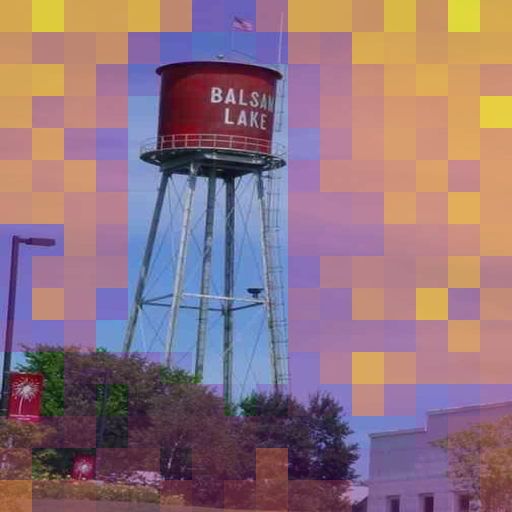}
        \hfill
        \includegraphics[width=0.48\textwidth]{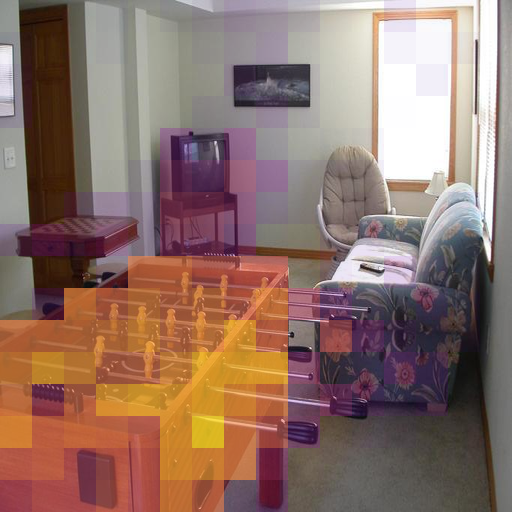}
        \caption{PCA}
        \label{fig:additional-ade20k-examples-pca}
    \end{subfigure}
    \hfill
    \begin{subfigure}[b]{0.23\textwidth}
        \centering
        \includegraphics[width=0.48\textwidth]{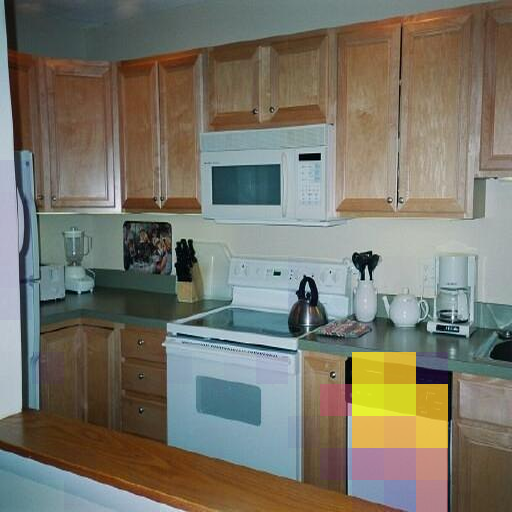}
        \hfill
        \includegraphics[width=0.48\textwidth]{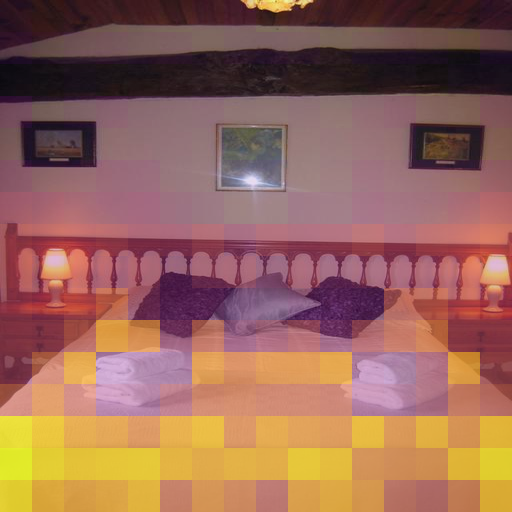}
        \includegraphics[width=0.48\textwidth]{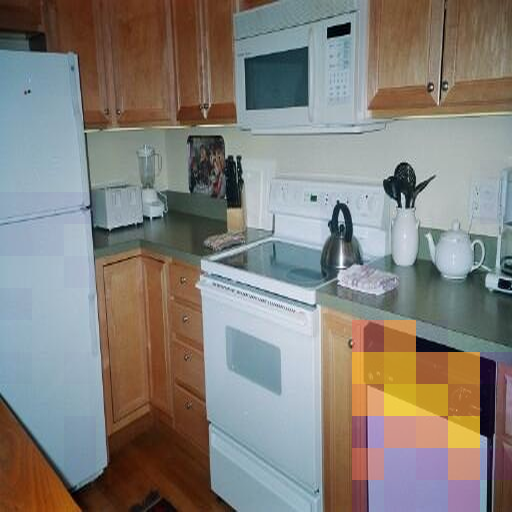}
        \hfill
        \includegraphics[width=0.48\textwidth]{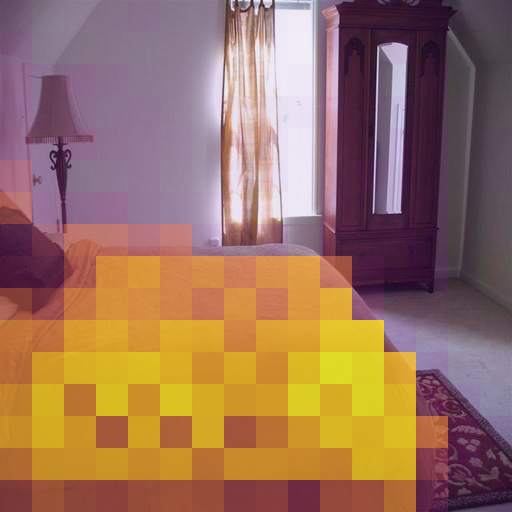}
        \includegraphics[width=0.48\textwidth]{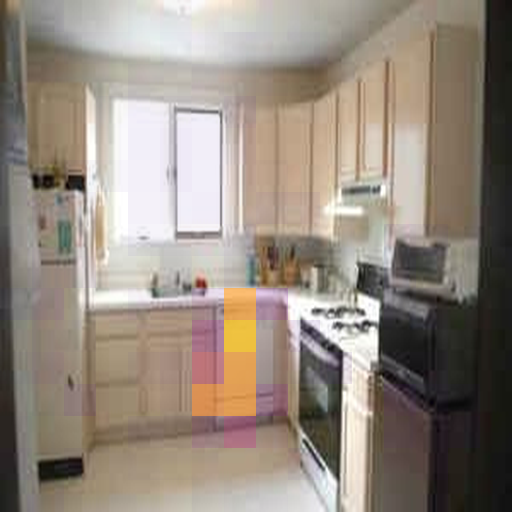}
        \hfill
        \includegraphics[width=0.48\textwidth]{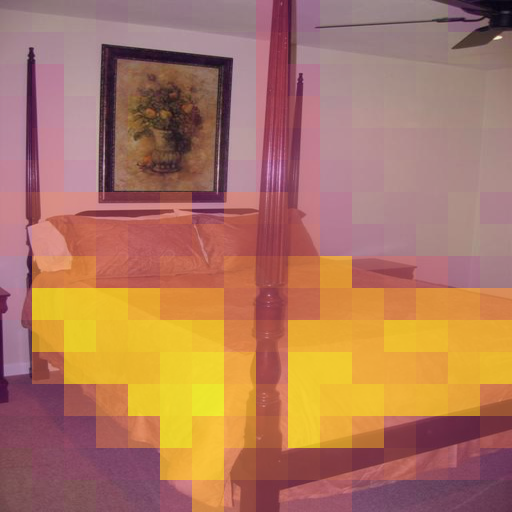}
        \includegraphics[width=0.48\textwidth]{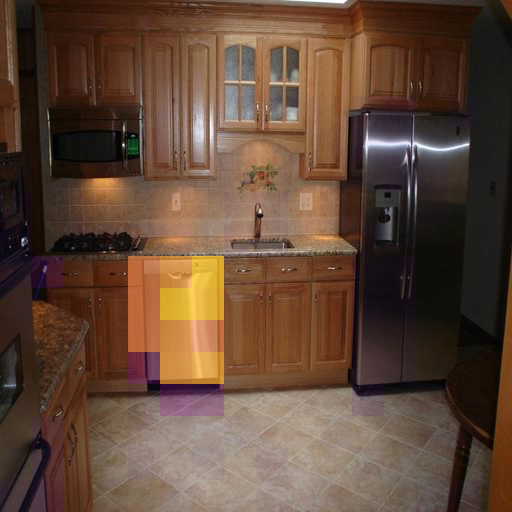}
        \hfill
        \includegraphics[width=0.48\textwidth]{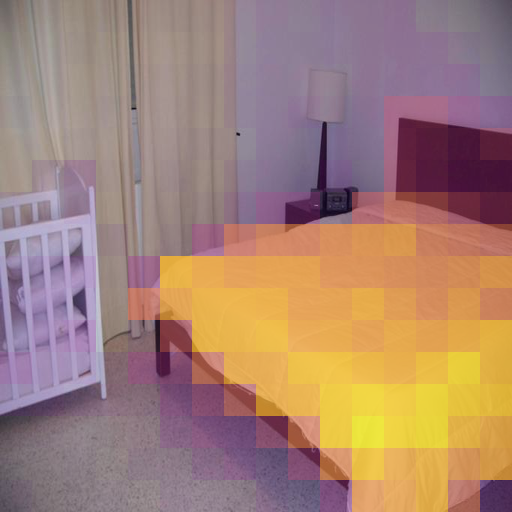}
        \caption{Vanilla SAE}
        \label{fig:additional-ade20k-examples-vanilla}
    \end{subfigure}
    \hfill
    \begin{subfigure}[b]{0.23\textwidth}
        \centering
        \includegraphics[width=0.48\textwidth]{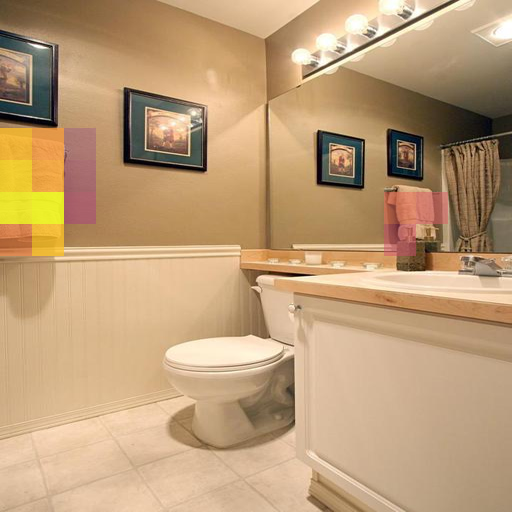}
        \hfill
        \includegraphics[width=0.48\textwidth]{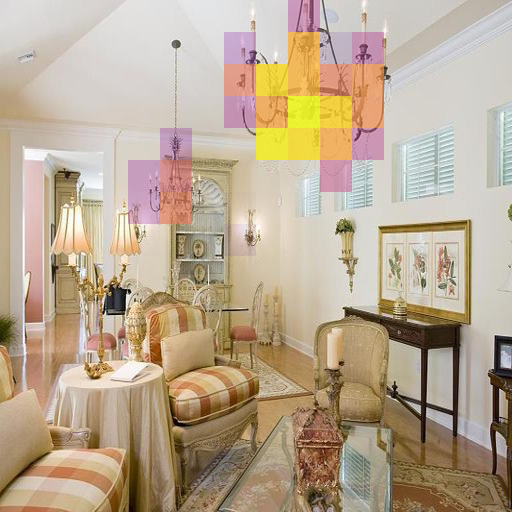}
        \includegraphics[width=0.48\textwidth]{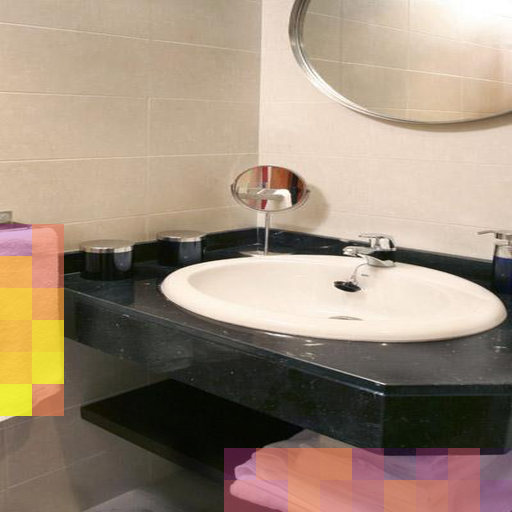}
        \hfill
        \includegraphics[width=0.48\textwidth]{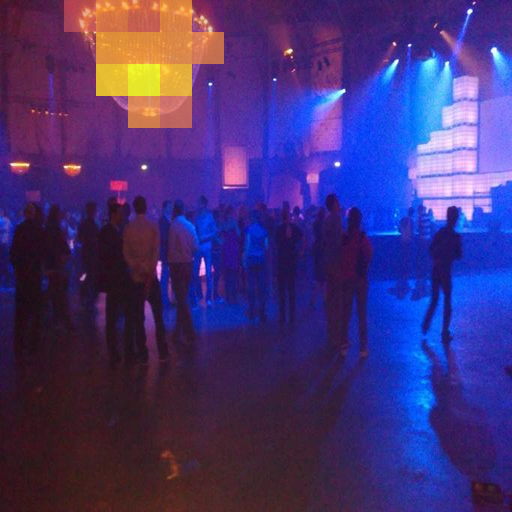}
        \includegraphics[width=0.48\textwidth]{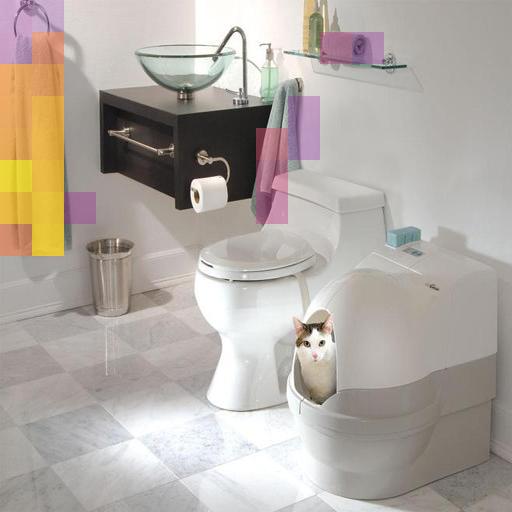}
        \hfill
        \includegraphics[width=0.48\textwidth]{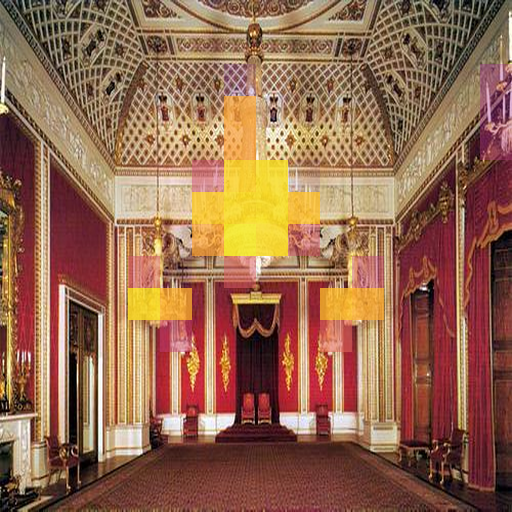}
        \includegraphics[width=0.48\textwidth]{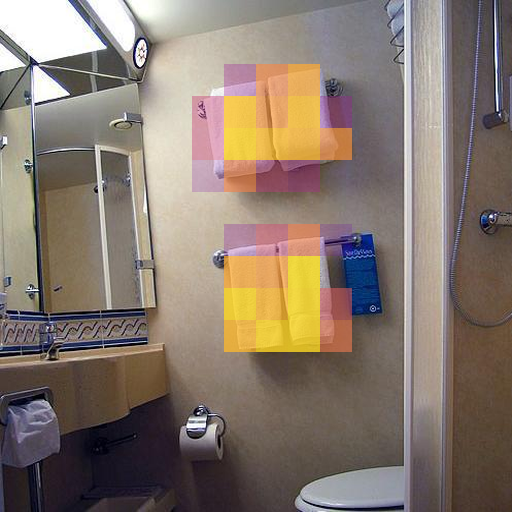}
        \hfill
        \includegraphics[width=0.48\textwidth]{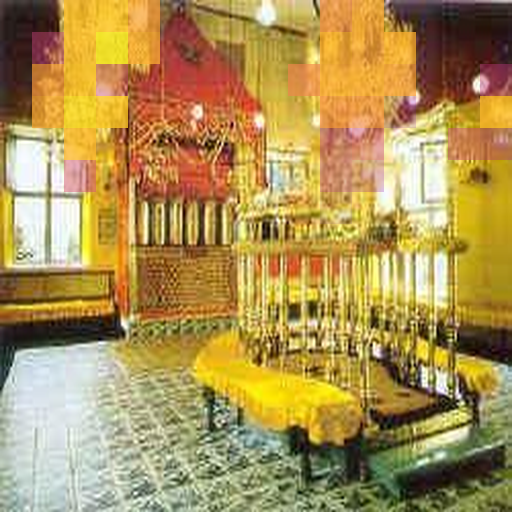}
        \caption{Matryoshka SAE}
        \label{fig:additional-ade20k-examples-matryoshka}
    \end{subfigure}
    \caption{  
    Some additional qualitative examples for \textbf{(a)} $k$-means: ``waterfall'' (left) and ``road'' (right), \textbf{(b)} PCA: ``sky'' (left) and ``pool table'' (right), \textbf{(c)} vanilla SAEs: ``dishwasher'' (left) and ``bed'' (right) and \textbf{(d)} Matryoshka SAEs: ``towel'' (left) and ``chandelier'' (right).
    }\label{fig:additional-ade20k-examples}
\end{figure*}

We show additional qualitative examples of both success and failure for ADE20K classes for all methods in \Cref{fig:additional-ade20k-examples}.

\section{Butterfly SAE Training}\label{app:butterfly-saes}

Our code contains instructions to reproduce our exact findings.

\paragraph{Butterfly Image Patch Activations}
\begin{itemize}
    \item{Patch activations for layers 22 and 24 of DINOv3 ViT-L/16}
    \item{384 patches per image, ratio-aware resized}
    \item{Only dorsal images of each subspecies pair used for calculating selectivity}
\end{itemize}

\paragraph{SAE Training.}
\begin{itemize}
    \item{Initialize $W_\text{enc}$ and $W_\text{dec}$ with Kaiming initialization \citep{kaiming2015init} and $b_\text{enc}$ and $b_\text{dec}$ with all zeros.}
    \item{Train Matryoshka TopK SAEs with Adam \citep{kingma2014adam} for \num{100}M examples with a batch size of \num{16384}.}
    \item{10 prefixes sampled for each training step}
    \item{Linearly warmup learning rate from \num{0} to the maximum over 500 steps, then use cosine decay to \num{0}.}
    \item{Normalize $W_\text{dec}$ to have unit norm columns at every step and remove gradients parallel to the columns to avoid interactions between Adam moments and normalization \citep{bricken2023monosemanticity}.}
    \item{Sweep both learning rate and $k$ for each setting, using the values in \Cref{tab:butterfly-hyperparams}.}
\end{itemize}

\begin{table}[t]
    \small
    \centering
    \begin{tabular}{ll}
    \toprule
    Hyperparam & Sweep Values \\
    \midrule
    Learning Rate & [\num{3e-4}, \num{1e-3}, \num{3e-3}]\\
    $k$ & [16, 32, 64, 128] \\
    \bottomrule
    \end{tabular}
    \caption{Sparse autoencoder hyperparameter sweep values for SAEs trained on \textit{Heliconius} images.}
    \label{tab:butterfly-hyperparams}
\end{table}

\section{Ablation Results}\label{app:ablation-results}

We report all metrics for the best SAE for each ViT size and layer in \Cref{tab:all-ablation-results}.

\paragraph{Model size.}\label{sec:ablation-size}
Model scale consistently correlates with downstream performance \citep{kaplan2020scaling,hoffmann2022training}.
Prior work proposes scaling laws for SAE reconstruction based on both SAE and foundation model size \citep{gao2025scaling} but does not investigate downstream performance.
We assess foundation model size effects by training SAEs on patch activations from DINOv3 ViT-S/16, ViT-B/16, and ViT-L/16 checkpoints.
The models vary in both total parameter count (\num{22}M, \num{86}M and \num{303}M, respectively) and the embedding dimension $d$ (\num{384}, \num{768}, and \num{1024}, respectively).
For each size, we keep the SAE width fixed at \num{16384}, train on ImageNet-1K \num{100}M patch activations and probe the SAEs for ADE20K features.

\begin{figure*}[t]
    \small
    \centering
    \begin{subfigure}[b]{0.48\textwidth}
        \includegraphics[width=\linewidth]{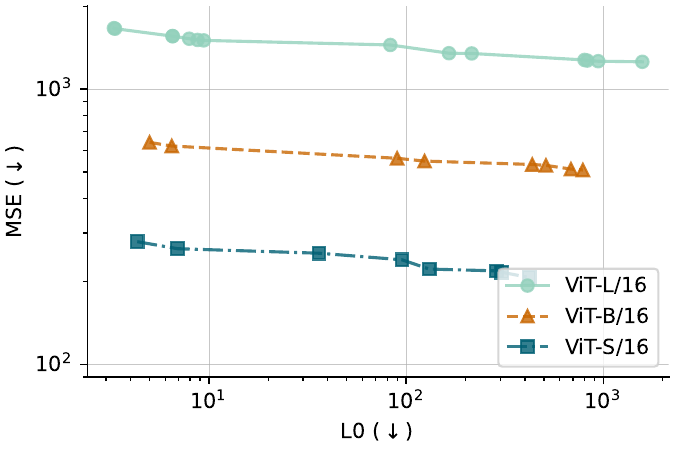}
    \end{subfigure}
    \hfill
    \begin{subfigure}[b]{0.48\textwidth}
        \centering
        \includegraphics[width=\linewidth]{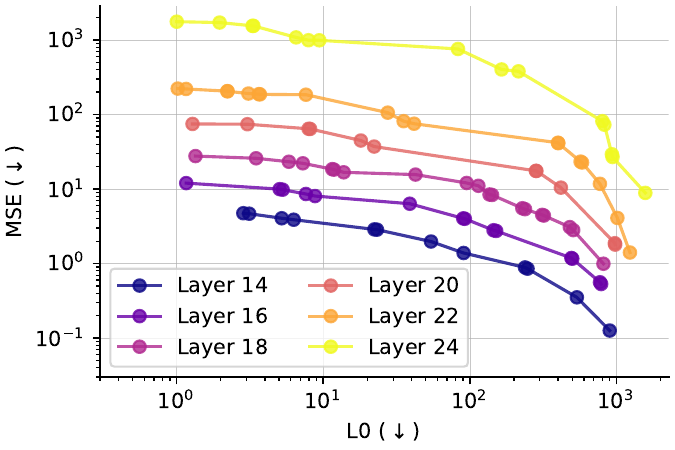}
    \end{subfigure}
    \caption{\textbf{(Left)} The reconstruction--sparsity tradeoff for SAEs trained on different sized ViTs. Larger backbones are harder to reconstruct at a given sparsity (ViT-L/16 $>$ ViT-B/16 $>$ ViT-S/16), though all improve as $L_0$ increases. Note that the $y$-axis is \textit{raw} rather than normalized MSE.
    \textbf{(Right)} Layer-wise reconstruction–sparsity trade-off (DINOv3 ViT-L/16) on ImageNet-1K. Later layers consistently incur higher MSE at matched sparsity, indicating greater representational complexity.
    }\label{fig:ablations-size-and-layer}
\end{figure*}

In \Cref{fig:ablations-size-and-layer} (left) we see that larger models are in fact harder to reconstruct (note that the $y$-axis is \textit{raw} rather than normalized MSE), consistent with prior work in scaling SAEs for language models \citep{gao2025scaling}.
Despite more challenging dictionary learning, we see in \Cref{app:layer-metrics} that SAEs applied to larger models both recover more concepts (coverage) and are more precise (purity).
All three sizes of the DINOv3 ViTs are distilled from a single, larger teacher model \citep{simeoni2025dinov3}; we conjecture then that all three checkpoints have learned the same set of concepts.
These results indicate that, at least for vision, strong foundation models in combination with SAEs can rediscover semantic concepts, which we view as a prerequisite for scientific discovery.

A potential confounder is total training FLOPs: prior work shows that additional SAE training compute (total FLOPs) leads to stronger SAEs as measured by the reconstruction--sparsity tradeoff \citep{gao2025scaling}.
Our SAEs trained on larger embedding dimensions $d$ use more compute during training; this could explain the stronger results for larger models.
However, our SAEs trained on larger models have \textit{worse} reconstruction--sparsity but \textit{better} downstream concept alignment metrics.

\paragraph{Layer depth.}\label{sec:ablations-layer}
Representations evolve across network depth: later layers in CNNs and ViTs tend to encode higher-level, more linearly separable semantics, which often improves transfer and dense alignment \citep{yosinski2014transfer,azizpour2014factors,caron2021dinov1}. 
However, recent results show that the final layer is not always optimal \citep{bolya2025perception}.
Intermediate layers can yield less entangled, more spatially localized features depending on the downstream task. 
We therefore evaluate six layers in the second half of the transformer for the three evaluated checkpoints: for the \num{24}-layer ViT-L/16 model, we use layers \num{14}, \num{16}, \num{18}, \num{20}, \num{22}, and \num{24} and for the \num{12}-layer ViT-S/16 and ViT-B/16 we use the last \num{6} layers \num{7}, \num{8}, \num{9}, \num{10}, \num{11}, and \num{12}.

\begin{figure*}[t]
    \centering
    \includegraphics[width=0.8\linewidth]{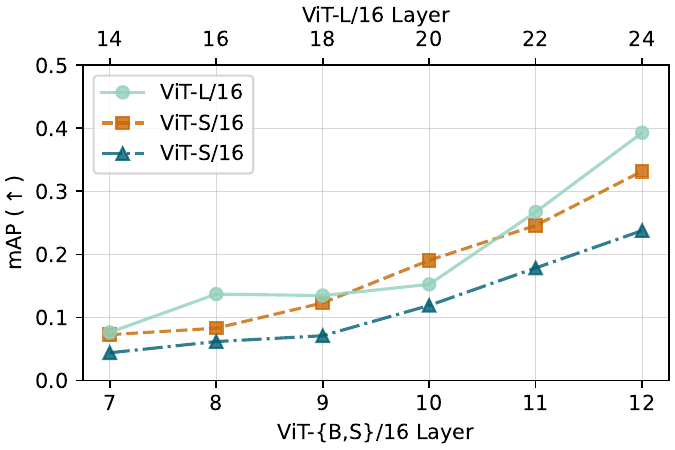}
    \caption{Matryoshka SAE probe quality as measured by mean average precision (mAP) for DINOv3 ViT-S/16, ViT-B/16 and ViT-L/16 checkpoints over different transformer layers. 
    \textbf{Takeaway:} Despite larger ViTs and later layers being harder to reconstruct at a given sparsity (see \Cref{fig:ablations-size-and-layer} for ViT size ablation), larger ViTs and later layers lead to better SAE concept alignment.
    }\label{fig:size-map}
\end{figure*}

In \Cref{fig:ablations-size-and-layer}, we find that later layers are harder to reconstruct (note that the $y$-axis is \textit{raw} rather than normalized MSE).
We also evaluate downstream metrics (mAP) for the best SAE at each layer for all three ViT checkpoints in \Cref{fig:size-map}.
Despite more challenging reconstruction, we find that later layers lead to better recovery for all three checkpoints.
Prior work in SAEs for language models finds that later layers are harder to reconstruct \text{except} for the final two layers \citep{gao2025scaling}.  
Language models use a linear embedding layer for next-token-prediction; the final layers must optimize for linear separability.
In contrast, DINOv3 uses multiple, loss-specific multilayer perceptrons with additional non-linearities for pre-training \citep{simeoni2025dinov3}; we conjecture this architectural difference leads to the difference in our results.

\section{Baseline Methods}\label{app:baselines}
\begin{figure}[t]
    \centering
    \small
    \small
    \centering
    \begin{subfigure}[b]{0.8\textwidth}
        \includegraphics[width=\linewidth]{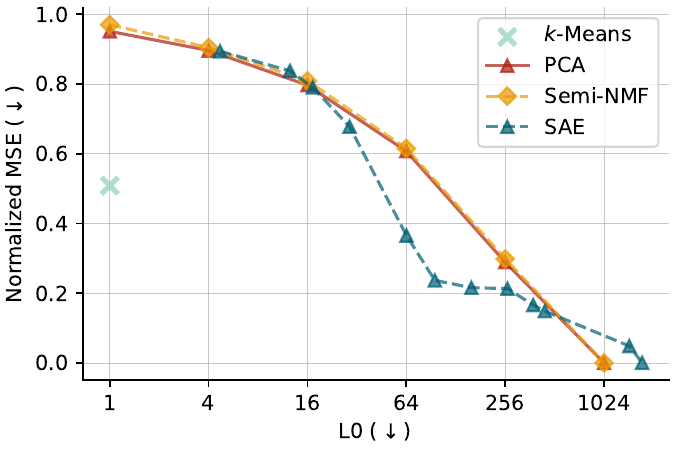}
    \end{subfigure}
    \caption{We compare $k$-means, PCA, SemiNMF and SAEs along the reconstruction--sparsity tradeoff for learning to decompose ViT patch activations. We use final layer activations from DINOv3 ViT-L/16 on ImageNet-1K; we fit all methods on the training split and measure normalized MSE (see \Cref{eq:nmse}) and L$_0$ on the validation split. For $k$-means, we ``reconstruct'' every patch with its nearest centroid; thus, L$_0$ is always $1$. For PCA, we sweep the number of components. We sweep $\lambda$ for the SAE and show the Pareto frontier of reconstruction--sparsity. \textbf{Takeaway:} Reconstruction--sparsity does not indicate an optimal method.
    }\label{fig:baseline-pareto-and-layerwise}
\end{figure}

We compare $k$-means, PCA, SemiNMF and SAEs along the reconstruction--sparsity tradeoff for learning to decompose ViT patch activations in \Cref{fig:baseline-pareto-and-layerwise}. While PCA and SemiNMF provide similar reconstruction-sparsity tradeoffs, we have demonstrated that they perform worse on downstream concept alignment.

\section{Additional Layer-Depth Ablations}\label{app:layer-metrics}

\begin{figure}[h]
    \centering
    \begin{subfigure}[b]{0.48\textwidth}
        \centering
        \includegraphics[width=\linewidth]{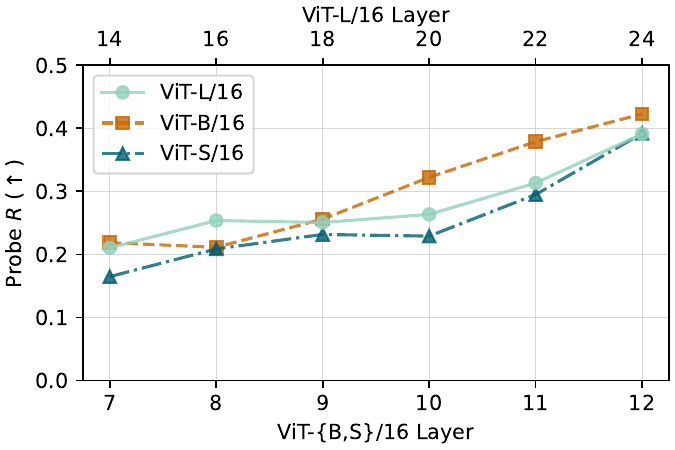}
        \caption{Probe $R$}
    \end{subfigure}
    \hfill
    \begin{subfigure}[b]{0.48\textwidth}
        \centering
        \includegraphics[width=\linewidth]{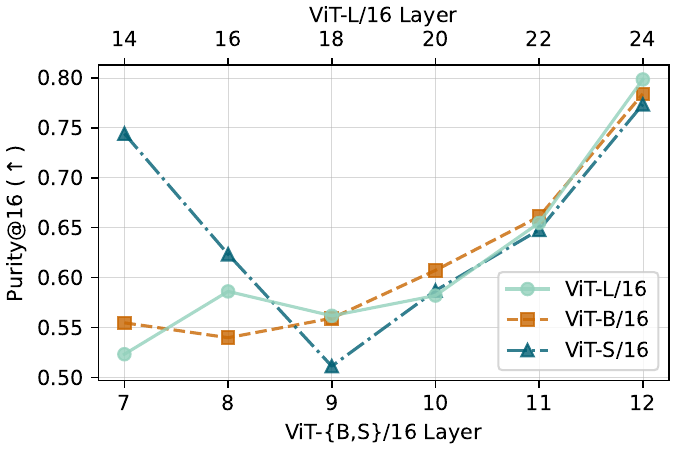}
        \caption{Purity@$k$}
    \end{subfigure}

    \begin{subfigure}[b]{0.48\textwidth}
        \centering
        \includegraphics[width=\linewidth]{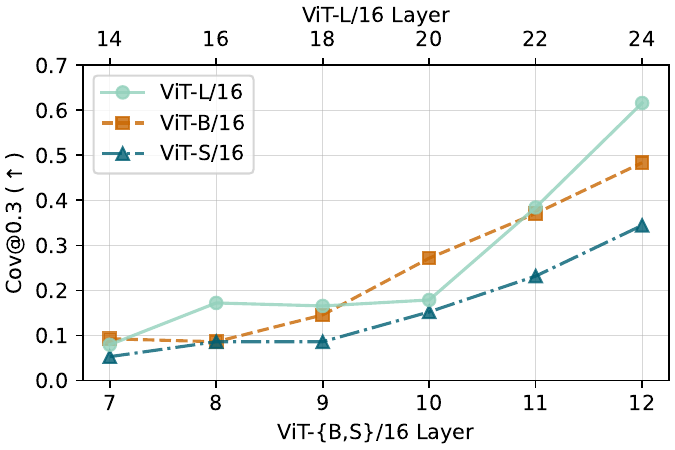}
        \caption{Coverage@$\tau$}
    \end{subfigure}
    \hfill
    \begin{subfigure}[b]{0.48\textwidth}
        \centering
        \includegraphics[width=\linewidth]{figures/dinov3_in1k_ade20k_layers_map.pdf}
        \caption{mAP (reproduced from \Cref{fig:size-map})}
    \end{subfigure}
    \caption{Layer-depth ablation for all four downstream metrics on ADE20K, for DINOv3 ViT-S/16, ViT-B/16, and ViT-L/16. The trend from \Cref{fig:size-map} (later layers improve concept alignment) holds across all metrics.}\label{fig:app-layer-metrics}
\end{figure}

\Cref{fig:size-map} reports mAP across layers; here we show the remaining metrics in \Cref{fig:app-layer-metrics} and \Cref{tab:all-ablation-results}.
All four metrics exhibit the same trend: later layers yield better concept alignment despite harder reconstruction, with larger models achieving slightly higher concept alignment in later layers.

\begin{table*}
    \small
    \centering
    \caption{Probe results for Matryoshka SAEs trained on ImageNet-1K and evaluated on ADE20K. For each layer in each size of ViT, we choose the best SAE based on training probe loss and report all dictionary learning and downstream metrics.}\label{tab:all-ablation-results}
    \begin{tabular}{lrrrrrrr}
    \toprule
    \multicolumn{2}{c}{Model} & \multicolumn{2}{c}{Dict. $\downarrow$} & \multicolumn{4}{c}{Downstream $\uparrow$} \\
    \cmidrule(lr){1-2} \cmidrule(lr){3-4} \cmidrule(lr){5-8}
    Size & Layer & NMSE & L$_0$ & Probe $R$ & mAP & Purity@$k$ & Coverage@$\tau$ \\
    \midrule
    ViT-S/16 & 7 & 0.182 & 70.9 & 0.165 & 0.044 & 0.744 & 0.053 \\
    ViT-S/16 & 8 & 0.143 & 135.1 & 0.209 & 0.062 & 0.623 & 0.086 \\
    ViT-S/16 & 9 & 0.133 & 159.8 & 0.232 & 0.071 & 0.511 & 0.086 \\
    ViT-S/16 & 10 & 0.299 & 84.1 & 0.251 & 0.099 & 0.626 & 0.093 \\
    ViT-S/16 & 11 & 0.438 & 23.6 & 0.310 & 0.160 & 0.645 & 0.205 \\
    ViT-S/16 & 12 & 0.294 & 95.1 & 0.434 & 0.203 & 0.742 & 0.245 \\
    \midrule
    ViT-B/16 & 7 & 0.572 & 19.5 & 0.216 & 0.055 & 0.548 & 0.066 \\
    ViT-B/16 & 8 & 0.497 & 35.4 & 0.240 & 0.061 & 0.516 & 0.073 \\
    ViT-B/16 & 9 & 0.571 & 10.7 & 0.265 & 0.109 & 0.613 & 0.126 \\
    ViT-B/16 & 10 & 0.431 & 20.1 & 0.322 & 0.191 & 0.607 & 0.272 \\
    ViT-B/16 & 11 & 0.271 & 47.5 & 0.379 & 0.246 & 0.661 & 0.371 \\
    ViT-B/16 & 12 & 0.359 & 89.9 & 0.441 & 0.312 & 0.812 & 0.430 \\
    \midrule
    ViT-L/16 & 14 & 0.252 & 90.7 & 0.210 & 0.076 & 0.523 & 0.079 \\
    ViT-L/16 & 16 & 0.336 & 92.4 & 0.266 & 0.117 & 0.575 & 0.152 \\
    ViT-L/16 & 18 & 0.659 & 11.8 & 0.261 & 0.090 & 0.533 & 0.099 \\
    ViT-L/16 & 20 & 0.471 & 22.2 & 0.263 & 0.152 & 0.582 & 0.179 \\
    ViT-L/16 & 22 & 0.380 & 35.3 & 0.313 & 0.267 & 0.655 & 0.384 \\
    ViT-L/16 & 24 & 0.703 & 6.5 & 0.403 & 0.319 & 0.809 & 0.450 \\
    \bottomrule
    \end{tabular}
\end{table*}

\section{Additional Butterfly Feature Images}\label{app:butterfly-images}
\begin{figure}[b]
    \centering
    \includegraphics[width=0.95\linewidth]{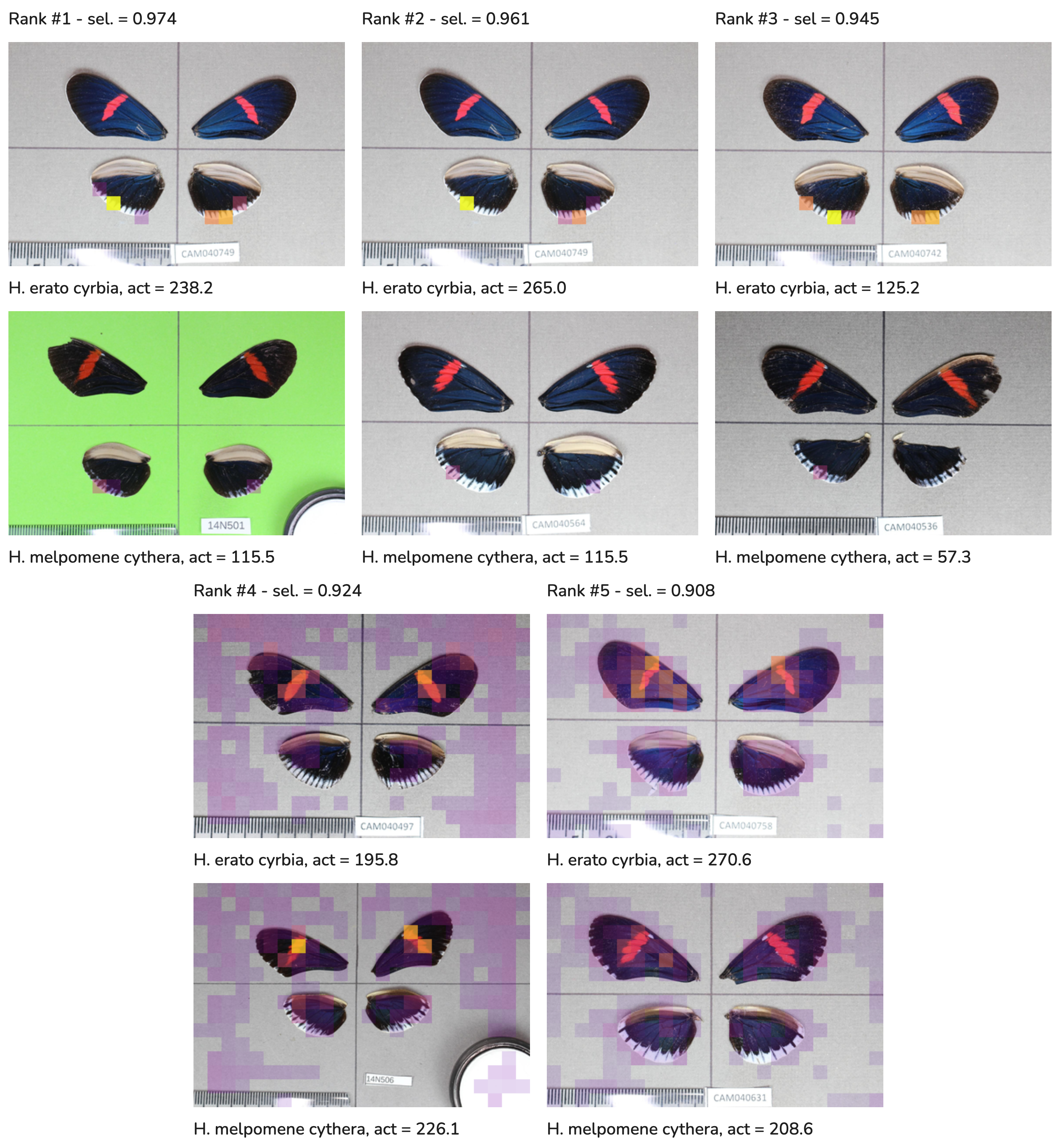}
    \caption{5 highest selectivity features for \textit{H. erato cyrbia}/\textit{H. melpomene cythera}. Features 1-3 correspond to hindwing fringe in \textit{H. erato cyrbia}, a known diagnostic trait. Features 4 and 5 may correspond to the shape of the forewing red patch, a known diagnostic trait, but these are not counted as corresponding to a known trait due to difficulty interpreting diffuse activation patterns.}
    \label{fig:cyrbia_cythera_top_5}
\end{figure}

\begin{figure}
    \centering
    \includegraphics[width=0.95\linewidth]{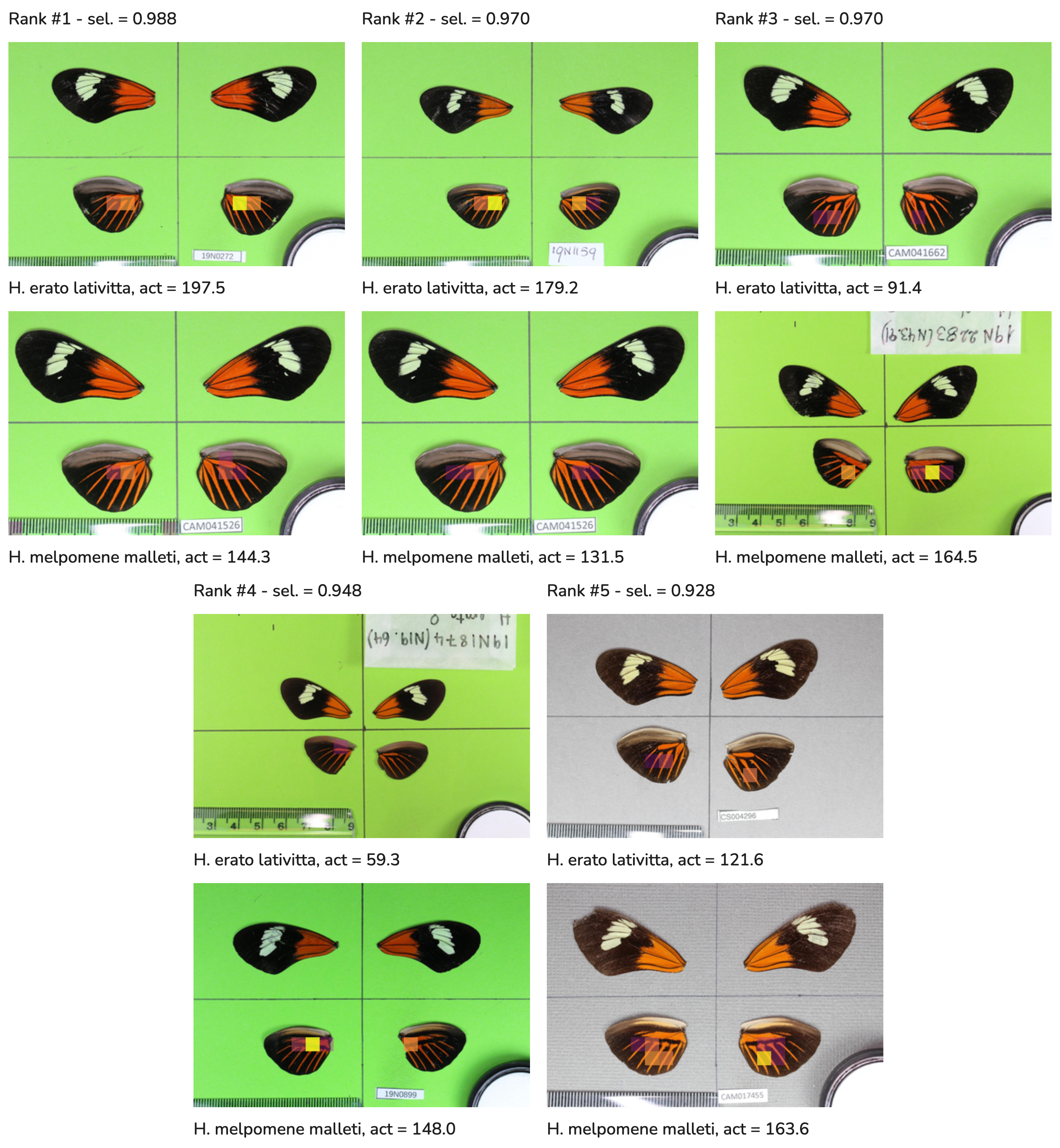}
    \caption{5 highest selectivity features for \textit{H. erato lativitta}/\textit{H. melpomene malleti}. All 5 features correspond to the black bar or striping on the hindwing, a known diagnostic trait in this mimic pair.}
    \label{fig:lativitta_malleti_top_5}
\end{figure}

\begin{figure}
    \centering
    \includegraphics[width=0.95\linewidth]{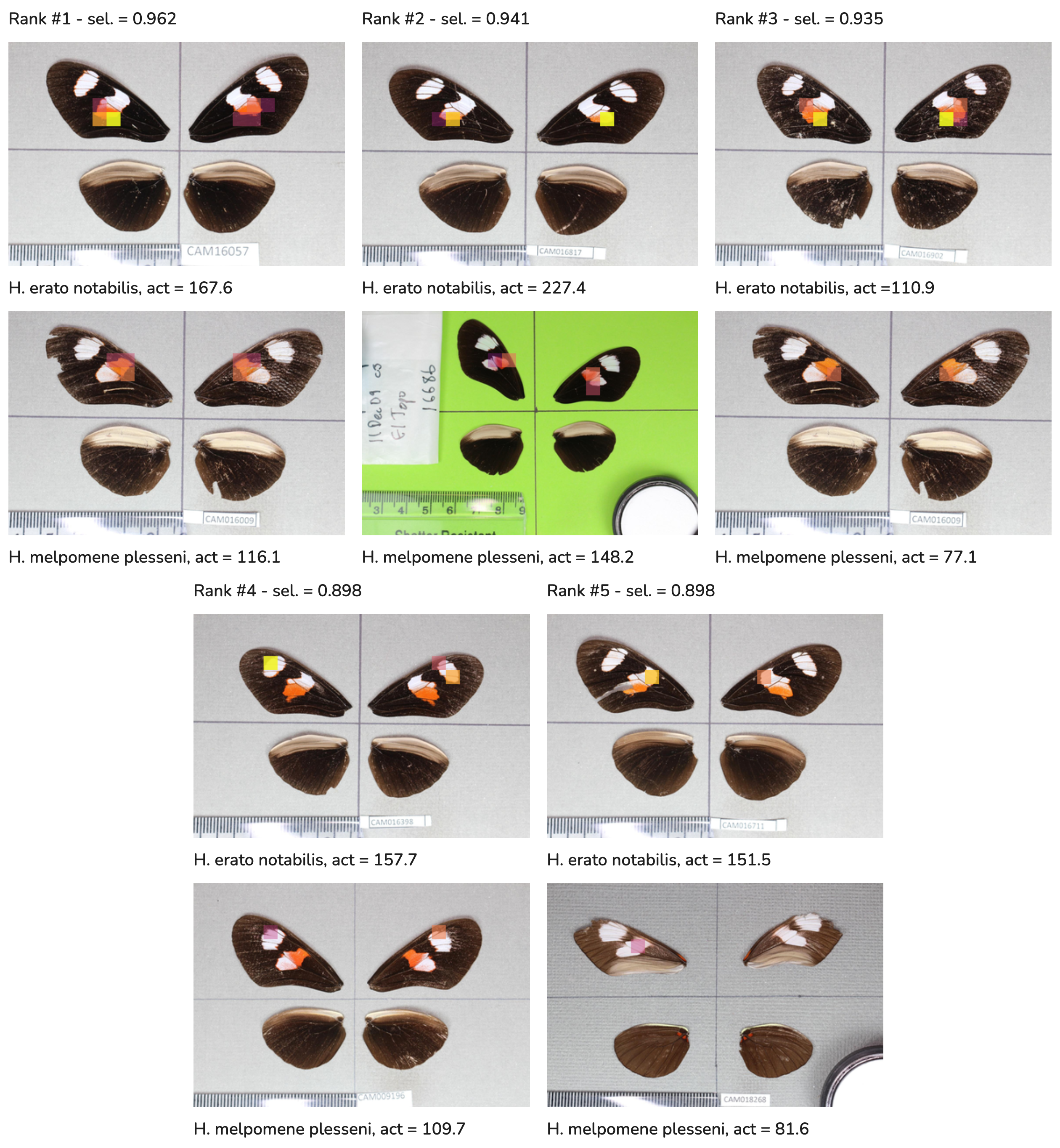}
    \caption{5 highest selectivity features for \textit{H. erato notabilis}/\textit{H. melpomene plesseni}. Features 1, 2, 3, and 5 correspond to the red forewing patch placement in \textit{H. erato notabilis}, a known diagnostic trait. Feature 4 corresponds to the upper forewing white patch in \textit{H. erato notabilis}. No diagnostic trait for this portion of the wing is well-documented in this mimic pair.}
    \label{fig:notabilis_plesseni_top_5}
\end{figure}

We provide the five highest-ranked features by selectivity across all pareto-optimal SAEs on layer 22 for each mimic pair, as well as their annotations: \textit{H. erato cyrbia} and \textit{H. melpomene cythera} (\Cref{fig:cyrbia_cythera_top_5}), \textit{H. erato lativitta} and \textit{H. melpomene malleti} (\Cref{fig:lativitta_malleti_top_5}), and \textit{H. erato notabilis} and \textit{H. melpomene plesseni} (\Cref{fig:notabilis_plesseni_top_5}).

\end{document}